\documentclass[twocolumn,10pt,letterpaper]{article}

\usepackage[letterpaper,textwidth=6.75in,textheight=9.25in,columnsep=0.25in]{geometry}
\usepackage{times}
\usepackage{microtype}
\usepackage[round]{natbib}

\usepackage{amsmath,amssymb,amsthm,mathtools}
\usepackage{url}
\usepackage{booktabs}
\usepackage{multirow}
\usepackage{graphicx}
\usepackage{algorithm}
\usepackage{algpseudocode}
\usepackage{xcolor}
\usepackage{subcaption}
\usepackage[normalem]{ulem}
\usepackage{cancel}
\usepackage[hidelinks]{hyperref}   % load last

\newcommand{\blfootnotetext}[1]{%
  \begingroup
  \renewcommand{\thefootnote}{}%
  \footnotetext{#1}%
  \endgroup
}

\theoremstyle{plain}
\newtheorem{proposition}{Proposition}
\newtheorem{lemma}{Lemma}

\theoremstyle{definition}

\newcommand{\R}{\mathbb{R}}
\newcommand{\E}{\mathbb{E}}
\newcommand{\Prob}{\mathbb{P}}

\newcommand{\Fcal}{\mathcal{F}}   % group-level predictor class F_g
\newcommand{\Xcal}{\mathcal{X}}   % feature space
\newcommand{\Ycal}{\mathcal{Y}}   % outcome space
\newcommand{\Zcal}{\mathcal{Z}}   % latent space
\newcommand{\Simplex}{\Delta_{G}} % probability simplex
\newcommand{\Wtwo}{\mathrm{W}_2}

\newcommand{\Loss}{\mathcal{L}}
\newcommand{\Ach}[1]{\Loss_{#1}^{\star}}   % optimal loss of group #1
\newcommand{\Reg}{\mathcal{R}}             % excess loss (not \Delta: clashes with the simplex)
\newcommand{\Risk}{R}
\newcommand{\Const}[1]{\Risk_{#1}^{\mathrm{const}}} % best constant predictor
\newcommand{\RefPt}[1]{\widetilde{\Risk}_{#1}}      % reference value used
\newcommand{\Est}[1]{\widehat{\Risk}_{#1}}          % cross-fitted estimate

\title{\bfseries Learning Across Heterogeneous Feature Spaces with Group Robustness}
\author{%
  Xenia Konti$^{*}$ \quad
  Arjun Tschand$^{*}$ \quad
  Chuan Hong \quad
  Nicoleta J.\ Economou-Zavlanos \quad
  Michael M.\ Zavlanos\\[0.6ex]
  Duke University%
}
\date{}

\begin{document}

\maketitle

\begingroup
\renewcommand{\thefootnote}{\fnsymbol{footnote}}
\footnotetext[1]{Equal contribution.}
\endgroup
\blfootnotetext{This work is supported in part by The Duke Endowment (TDE) under grant \#7262-SP and by the Onassis Foundation under award \#ZT033-1/2023-2024.}
\setcounter{footnote}{0}

\begin{abstract}
In clinical applications, patients undergo different combinations of tests, and machine learning models must therefore ensure accurate predictions across heterogeneous feature spaces. Current approaches upweight the patients with the highest loss, without accounting for the limited predictive information in their features preventing their loss from going any lower. Moreover, when patients' feature spaces are fully disjoint, current methods reduce to training independent models, preventing knowledge transfer across them. We propose a framework for group robust learning across heterogeneous feature spaces, where groups are patients sharing a feature space. Since a high loss may simply reflect uninformative features, we instead optimize for worst-group \emph{excess loss}, the gap between a group's loss and an estimate of its achievable optimum. To allow collaboration between groups, we use group-specific encoders mapping into a shared latent space in which learnable Gaussian anchors cluster patients by outcome. On tabular and imaging medical datasets, including fully disjoint feature settings, our framework outperforms state-of-the-art methods when groups differ in their optimal losses or cannot train a good model on their own, reducing worst-group loss by up to 79\% and excess loss by up to 95\% with up to 18$\times$ fewer parameters.
\end{abstract}

\section{Introduction}
\label{sec:intro}

\begin{figure}[t]
  \centering
  \includegraphics[width=\linewidth]{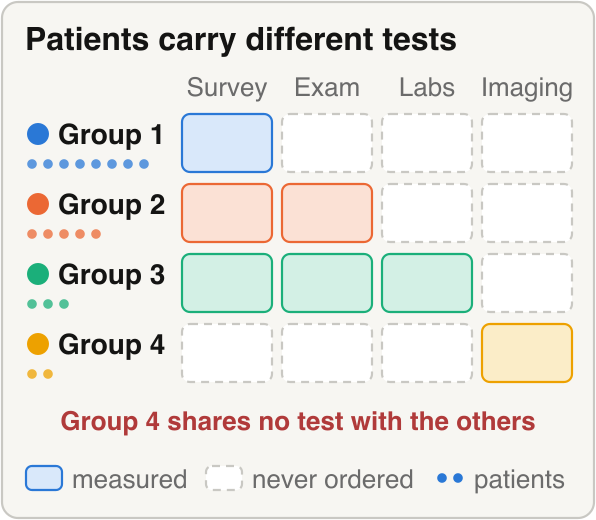}
  \caption{Patients receive different combinations of tests. Patients who share the same combination form a group, and groups may have disjoint feature spaces.}
  \label{fig:problem}
\end{figure}

Supervised learning typically assumes that all observations in a dataset share a common feature space \citep{vapnik1998statistical,hastie2009elements,shalev2014understanding}. In clinical settings, however, this assumption rarely holds. Modern care offers a wide variety of medical tests, from imaging and laboratory panels to clinical notes and genomics \citep{acosta2022multimodal,kline2022multimodal,lipkova2022artificial}, and there is no fixed protocol for which tests should be ordered in each case \citep{osullivan2018practice,li2021variation,wabe2021atlas}. Even for the same test, differences in equipment or in the people performing it can lead to different levels of fidelity \citep{han2006reliability,glocker2019scanner,castro2020causality}. As a result, patients receive different combinations of tests, which in machine learning terms translates to different feature spaces. We refer to patients who share a feature space as a \emph{group}. Maintaining a separate model for each group can increase the monitoring and maintenance costs of a clinical deployment \citep{liao2022governance,van2023implementation,davis2024sustainable}. Instead, a hospital needs one model that uses all the tests a patient received and performs well for all patients across heterogeneous feature spaces.

\begin{figure*}[t]
  \centering
  \begin{subfigure}[t]{0.61\textwidth}
    \centering
    \includegraphics[width=\linewidth]{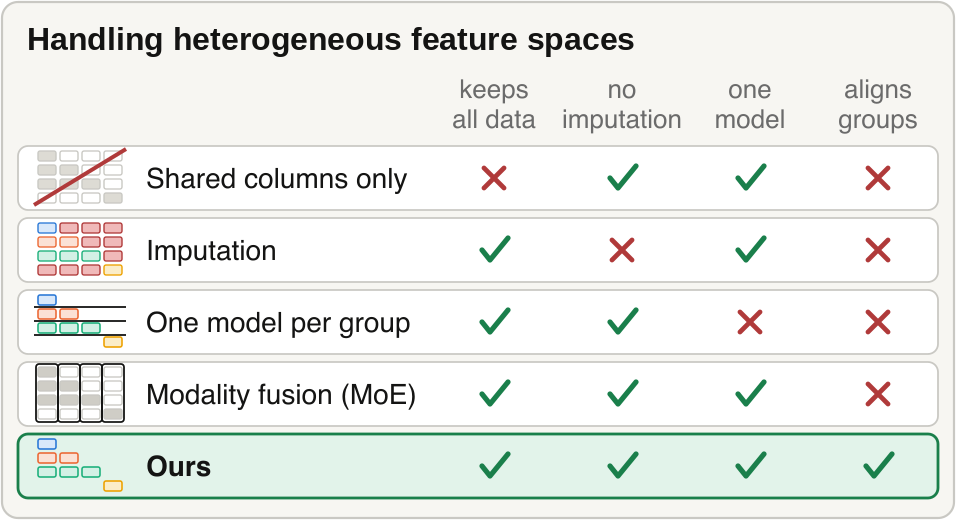}
    \caption{{Our framework uses all the available information, does not impute missing data, trains a single model for all, and allows collaboration across groups even when feature spaces are fully disjoint.}}
    \label{fig:methods}
  \end{subfigure}
  \hfill
  \begin{subfigure}[t]{0.37\textwidth}
    \centering
    \includegraphics[width=\linewidth]{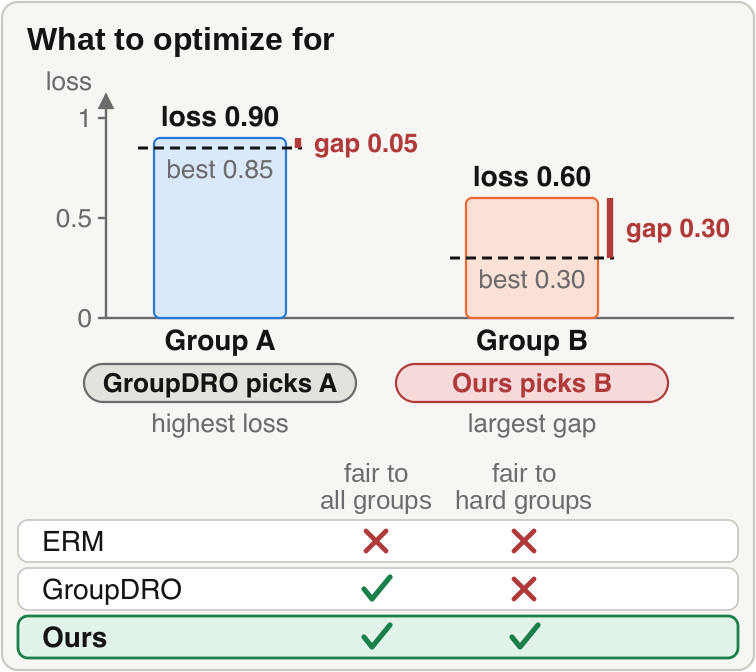}
    \caption{Raw-loss and excess-loss criteria can prioritize different groups.}
    \label{fig:objective}
  \end{subfigure}
  \caption{Learning across heterogeneous feature spaces: how to share information across groups (a) and what to optimize (b).}
  \label{fig:overview}
\end{figure*}

To train such a model, existing approaches first construct a common feature space, either by removing the features that are not shared by all patients \citep{rubin1976inference,schafer2002missing,little2019statistical} or by imputing the missing information \citep{vanbuuren2011mice,yoon2018gain,yuan2024remind}. The former can leave a large part of the available information unused, and the latter may introduce estimation error. Another line of work 
keeps all the available information and focuses on transferring knowledge across groups, by formulating the problem as a mixture of experts. 
For each test recorded in the dataset, an expert is trained, and for each group, a fusion function combines the experts of the tests that its patients received \citep{wang2023shaspec,han2024fusemoe,yun2024flexmoe,wu2026remind}. Groups that share a test thus transfer knowledge through its expert. However, when feature spaces are disjoint, no such collaboration is guaranteed.

Independently of how the feature spaces are handled, the resulting models are also affected by what they are trained to optimize. A common choice is empirical risk minimization \citep{vapnik1991principles}, which optimizes the average loss over all patients \citep{yun2024flexmoe,han2024fusemoe,yang2026poe}. However, groups are rarely equally represented in clinical datasets, and how often each group appears can change over time \citep{osullivan2018practice,quinonero2009dataset,koh2021wilds}. Average-loss training gives more weight to larger groups and neglects the rest. Recent work addresses this issue with Group Distributionally Robust Optimization \citep{oren2019distributionally,sagawa2020distributionally,duchi2021learning}, which optimizes the loss of the worst-performing group instead \citep{wu2026remind}. Still, these methods compare groups by their raw loss, without accounting for the fact that, depending on the information contained in their features, groups differ in the loss they can achieve. As a result, worst-group training may upweight groups that are simply hard to predict, rather than groups that the model has not been properly optimized for. 

In this work, we introduce a framework that addresses both problems: a single model across heterogeneous feature spaces, trained for the groups that can still improve. Each group has its own encoder, which maps its features into a latent space shared by all groups, and a single head predicts the outcome from this latent space. To allow collaboration between groups even when their feature spaces are disjoint, we organize the latent space around class-based Gaussian anchors: an alignment loss pulls the embeddings of each class, in every group, towards its anchor, and a separation loss encourages the anchors to be distinct. Since groups have different optimal losses, we train the model to minimize the worst-group excess loss \citep{agarwal2022minimax}, with the optimal losses estimated by nested cross-validation.

Our contributions are the following. We formalize group-robust learning across heterogeneous feature spaces, motivate its two design choices theoretically (Propositions~\ref{prop:order} and~\ref{prop:separate}), and propose an algorithm that estimates the optimal losses and trains the model. We evaluate our framework on three medical datasets against state-of-the-art methods, and ablation and synthetic experiments show when each component is useful. Our method outperforms these methods when groups differ in their optimal losses and when groups cannot train a good model on their own but benefit from collaborating, with up to 79\% lower worst-group loss and up to $18\times$ fewer parameters.

\providecommand{\newtxt}[1]{\textcolor{magenta}{#1}}
\providecommand{\deltxt}[1]{\textcolor{magenta}{\sout{#1}}}

\section{Problem setup}
\label{sec:setup}

We consider a classification task with a dataset
$\mathcal{D} = \{(x_i, y_i)\}_{i=1}^{N}$ of $N$ patients, where
$y_i \in \Ycal$ is the label and $|\Ycal| = C$.
The patients are not described by a common set of covariates. Instead,
each patient carries a subset of $m$ possible tests
$\{M_1, \dots, M_m\}$, where a test is any recorded source of
information, such as a laboratory panel, a clinical note or a mammographic
view, and the same test recorded with different equipment counts as a different test. \newline
We assume that there are $G$ distinct combinations of these tests that appear in the
dataset, and we define each combination as a unique feature space, denoted as $\Xcal_1, \dots, \Xcal_G$. Patients who share a feature space form a
\emph{group}, and we denote by $\mathcal{D}_g$ the training patients of group $g$. \newline
Each patient in the dataset is generated by first sampling the group $g$
they belong to with probability $p_g > 0$, $\sum_g p_g = 1$, and then drawing their covariates and outcome $(x, y)$ from a
group-specific distribution $\Prob^g_{X,Y}$ on $\Xcal_g \times \Ycal$. We call $p = (p_1, \dots, p_G)$ the
\emph{group distribution}, which lies in the probability simplex
$\Simplex := \{\lambda \in \R^G : \lambda_g \ge 0,\ \sum_g \lambda_g = 1\}$.
The
data therefore follow the mixture
\begin{equation}
  \Prob_{X,Y} \;=\; \sum_{g=1}^{G} p_g\, \Prob^g_{X,Y}.
  \label{eq:mixture}
\end{equation}

Our goal is to learn, for a given nonnegative loss $\Loss$, a single model that predicts the
outcome well for every group. Since patients do not share a common feature space, we represent this model as a set of group predictors $f = (f_1, \dots, f_G)$, where
$f_g : \Xcal_g \to \R^C$ outputs class scores for the patients of group
$g$. These predictors are parameterized jointly as seen in Section~\ref{sec:model}. The expected loss of group $g$ is
\begin{equation}
  \Loss_g(f_g) \;:=\; \E_{(x,y) \sim \Prob^g_{X,Y}}\big[\Loss(f_g; x, y)\big].
  \label{eq:grouploss}
\end{equation}
We later refer to it as the task loss for each group. \newline 
Given the data-generating process \eqref{eq:mixture}, typical machine learning
techniques rely on empirical risk minimization (ERM) \citep{vapnik1991principles}, which minimizes the average loss over all patients, i.e., the group losses weighted by the group distribution:
\begin{equation}
  \min_{f}\ \sum_{g=1}^{G} p_g\, \Loss_g(f_g).
  \label{eq:erm}
\end{equation}

\subsection{Robustness to group shift}
\label{sec:dro}
The ERM problem \eqref{eq:erm} weights each group by the group distribution
$p$ of the training data. However, $p$ reflects how often each combination of tests happened to be ordered when the data were collected, and it may change as clinical practice evolves. Moreover, even if $p$ remains the same, a low average loss does not imply that the model performs well
on every group, particularly on those that are underrepresented in the training data. To promote equal performance for all groups, models are trained using Group Distributionally Robust Optimization (Group DRO) \citep{oren2019distributionally,sagawa2020distributionally,duchi2021learning}. \newline
Group DRO learns a model that performs best under the worst-case mixture of
the group distributions. Each $\lambda \in \Simplex$ represents a possible
group distribution at deployment, under which the data follow the mixture
$\sum_g \lambda_g \Prob^g_{X,Y}$ and the expected loss of $f$ is
$\sum_g \lambda_g \Loss_g(f_g)$. Rather than fixing the weights to the
training distribution $p$, as in \eqref{eq:erm}, group DRO minimizes this
loss under the worst-case $\lambda$:
\begin{equation}
  \min_{f}\ \max_{\lambda \in \Simplex}\ \sum_{g=1}^{G} \lambda_g\,
  \Loss_g(f_g).
  \label{eq:minimax}
\end{equation}
The maximum over the simplex is attained at a vertex, which means that Group DRO optimizes the loss of the worst-performing group.

\subsection{From loss to excess loss}
\label{sec:regret}

\begin{figure*}[t]
  \centering
  \includegraphics[width=\textwidth]{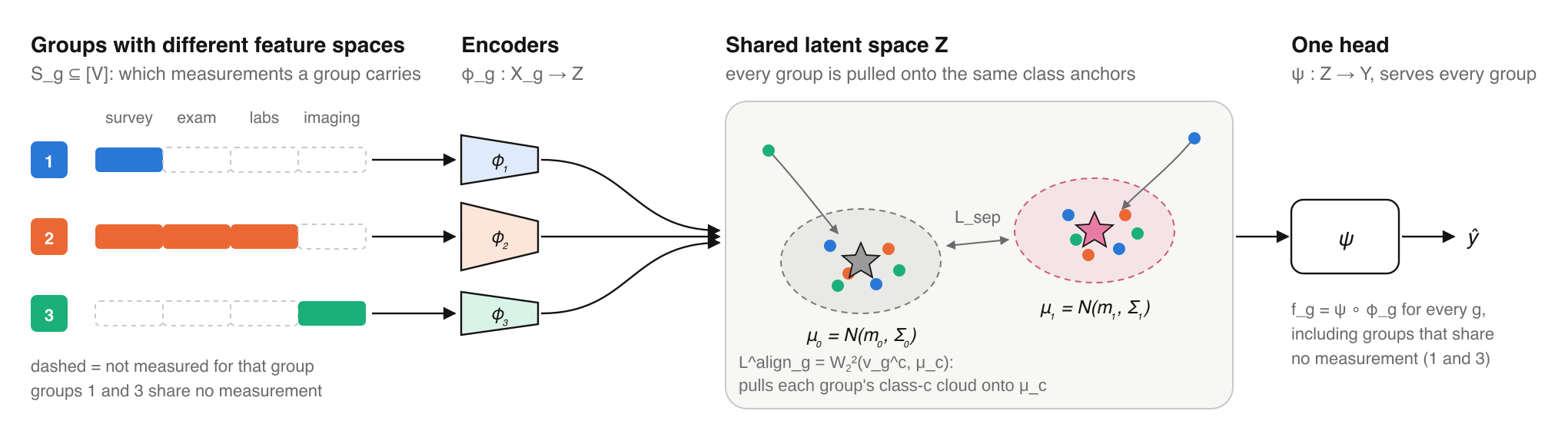}
  \caption{Overview of the model. Each group has its own encoder $\phi_g$ into a latent space $\Zcal$ shared by all groups, where Gaussian anchors $\mu_c$ pull the embeddings of each class toward the same region for every group. A single head $\psi$ serves all groups, including groups 1 and 3, which share no test.}
  \label{fig:method}
\end{figure*}

Group DRO compares groups by their absolute loss, which is informative only when groups have equal achievable losses. In the heterogeneous feature space problem, however, this is not always the case, since groups carry different information about the outcome.

To make this precise, let $\Fcal_g$ denote the class of predictors on
$\Xcal_g$ that our model can represent (specified in
Section~\ref{sec:model}), and define the optimal loss of group $g$ as
\begin{equation}
  \Ach{g} \;:=\; \inf_{f_g \in \Fcal_g} \Loss_g(f_g).
  \label{eq:optloss}
\end{equation}
Even two groups drawn
from the same population can have different optimal losses, solely because
they record different tests, as shown in the following proposition.
\begin{proposition}[More tests, lower optimal loss]
\label{prop:order}
Let groups $g$ and $h$ describe the same population, where $h$ records every
test of $g$ and additional ones, i.e.\ $\Xcal_h = \Xcal_g \times \Xcal'$
and $x_h = (x_g, x')$. If the model class of $h$ can ignore the extra
tests $x'$, then $\Ach{h} \le \Ach{g}$.
\end{proposition}
The proof is given in Appendix~\ref{app:regret}. \newline
In general, groups differ in
more than their tests. They may also differ in their patient populations or in how noisy their outcome is, which can also lead to different optimal losses. We therefore measure each group against its own optimal loss, and define the
\emph{excess loss} \citep{savage1951theory,agarwal2022minimax}
\begin{equation}
  \Reg_g(f_g) \;:=\; \Loss_g(f_g) - \Ach{g} \;\ge\; 0 .
  \label{eq:regret}
\end{equation}
An excess loss of zero means the same thing in every group: nothing further
is achievable within the model class. Under ERM
this change has no effect, since replacing $\Loss_g$ by $\Reg_g$ in
\eqref{eq:erm} only shifts the objective by the constant
$\sum_g p_g \Ach{g}$. However, it matters once the weights are chosen adversarially. In this work we adopt the minimax excess-loss framework \citep{agarwal2022minimax}:
\begin{equation}
  \min_{f}\ \max_{\lambda \in \Simplex}\ \sum_{g=1}^{G} \lambda_g
  \big(\Loss_g(f_g) - \Ach{g}\big).
  \label{eq:excessdro}
\end{equation}
\subsection{Model class and final objective}
\label{sec:model}

In this section we specify the form of the group predictors $f_g$. Each
predictor consists of two parts.
\begin{enumerate}
    \item A group-specific encoder
$\phi_g \in \Phi_g$ that maps the feature space of group $g$ into a $k$-dimensional latent space $\Zcal = \R^{k}$ common to all groups, $\phi_g : \Xcal_g \to \Zcal$
\item A head $\psi \in \Psi$, shared by
all groups that maps this latent space to class scores, $\psi : \Zcal \to \R^C$
\end{enumerate}
Thus, each group uses a model of the form $f_g = \psi \circ \phi_g$, with $f_g \in \Fcal_g = \{\psi \circ \phi_g : \psi \in \Psi,\ \phi_g \in \Phi_g\}$ (Figure~\ref{fig:method}), and groups may share information only through the common head $\psi$ and the common latent space $\Zcal$. However, collaboration between groups is still not guaranteed. Without constraints on the latent space, each encoder can map its group to a separate region of $\Zcal$, where the shared head can behave differently and make independent predictions for each group, as the following proposition shows. 
\begin{proposition}[A shared head does not make groups collaborate]
\label{prop:separate}
Assume that the encoders are bounded and the class $\Phi_g$ of each group is closed under adding a constant vector to its output, so that each group can place its representation anywhere in $\Zcal$, and that the head can be any function from $\Zcal$ to $\R^C$. Then, for any models $\psi_g\circ\phi_g$ of the groups, each with its own head $\psi_g$, there are encoders $\tilde\phi_g$ and a single shared head $\tilde\psi$ such that $\tilde\psi\circ\tilde\phi_g=\psi_g\circ\phi_g$ for every group $g$.
\end{proposition}
The proof is given in Appendix~\ref{app:separate}.
\newline
This does not limit what each group can achieve, since an independent model for each group can still perform well. However, it has two drawbacks. First, a group with few patients may not be able to learn a good encoder from its own data, and nothing encourages it to use what the other groups learn. Second, the latent space has no structure: the groups lie in unrelated regions, so the representations are hard to compare across groups, which limits their use beyond the prediction itself (Appendix~\ref{app:latent_space}).
\newline
To encourage sharing across groups and to give the latent space a structure, we regularize their representation in the latent space. Following work that couples separately trained
encoders through shared class prototypes \citep{snell2017prototypical,tan2022fedproto,rakotomamonjy2023flic}, we organize the latent space around the classes we aim to predict. For each class $c$, we learn a Gaussian anchor $\mu_c = \mathcal{N}(m_c, \Sigma_c)$ with diagonal covariance $\Sigma_c$, that serves as a common target for the patients of class $c$ in every group. 
Let $\nu_g^{c}$ denote the distribution of $\phi_g(X)$ for patients in
group $g$ with outcome $c$, and let $\tilde\nu_g^c$ be the diagonal Gaussian with the same mean and coordinate-wise variances. Using the 2-Wasserstein distance
\citep{villani2009optimal,peyre2019computational}, we define the alignment loss of group $g$ as the average squared
distance between $\tilde\nu_g^c$ and the corresponding anchor $\mu_c$:
\begin{equation}
  \Loss^{\mathrm{align}}_g
  \;=\; \frac{1}{C}\sum_{c=1}^{C} {\Wtwo^2\big(\tilde\nu_g^{c},\, \mu_c\big)}.
  \label{eq:align}
\end{equation}
This loss clusters patients of the same class together in the latent space, and it lower-bounds the Wasserstein loss between the full distributions $\nu_g^c$ and the anchors (Lemma~\ref{lem:lower}). However, to discourage class collapse, we also define a separation loss:
\begin{equation}
  \Loss^{\mathrm{sep}}
  \;=\; \frac{1}{C}\sum_{c=1}^{C} \E_{z \sim \mu_c}\big[\Loss(\psi; z, c)\big].
  \label{eq:sep}
\end{equation}
This loss encourages the anchors to be distinct enough for the shared head to correctly classify samples drawn from each of them. Details on the anchors and the mathematical formulation of the latent space losses are found in Appendix~\ref{app:losses}. 

Combining the excess-loss group DRO problem \eqref{eq:excessdro} with the
alignment and separation losses gives the problem formulation
\begin{equation}
\begin{split}
  \min_{\substack{\phi_1, \dots, \phi_G,\ \psi \\ \mu_1, \dots, \mu_C}}
  \Bigg\{ & \max_{\lambda \in \Simplex}\ \sum_{g=1}^{G} \lambda_g
  \big(\Loss_g(\psi \circ \phi_g) - \Ach{g} +\alpha_{\mathrm{align}}\, \Loss^{\mathrm{align}}_g\big) \\
  & + \alpha_{\mathrm{sep}}\, \Loss^{\mathrm{sep}} \Bigg\}, \text{ where } \alpha_{\mathrm{align}},\alpha_{\mathrm{sep}}\ge0 
\end{split}
  \label{eq:objective}
\end{equation}

\providecommand{\newtxt}[1]{\textcolor{magenta}{#1}}
\providecommand{\deltxt}[1]{\textcolor{magenta}{\sout{#1}}}

\section{Algorithm}
\label{sec:algorithm}

\subsection{Estimation of Optimal Losses}
\label{sec:floors}

The optimal loss $\Ach{g}$ in \eqref{eq:optloss} is the best loss that the model class can achieve for group $g$, and it is not known a priori. In practice, we replace it with an estimate $\RefPt{g}$, which we call the reference of group $g$, using the training data. To do so, we propose a nested cross-validation method as summarized in Algorithm~\ref{alg:floors}.
\begin{algorithm}[t]\small
\caption{Nested Cross-Validation Algorithm for estimating references}
\label{alg:floors}
\begin{algorithmic}[1]
\Require Training dataset $\mathcal{D}_g$ of every group $g$, set of candidate architectures $\mathcal{A}$, number of outer folds $K$ and inner folds $K_{\mathrm{in}}$, number of bootstrap resamples $N_{\mathrm{boot}}$, error level $\beta$
\Function{SelectPredictor}{$\mathcal{S}$} \Comment{inner level}
    \State Split $\mathcal{S}$ into $K_{\mathrm{in}}$ folds $V_1, \cdots, V_{K_{\mathrm{in}}}$ \label{alg:isplit}
    \For{each architecture $a \in \mathcal{A}$}
        \For{$j = 1$ to $K_{\mathrm{in}}$}
            \State Train $a$ on $\mathcal{S} \setminus V_j$ to obtain $\hat{f}_{a,j}$ \label{alg:itrain}
            \State $\ell_{a,j} \gets \{\Loss(\hat{f}_{a,j}; x, y) : (x, y) \in V_j\}$ \label{alg:iloss}
        \EndFor
    \EndFor
    \State \Return $\arg\min_{a \in \mathcal{A}}$ mean of $(\ell_{a,1}, \cdots, \ell_{a,K_{\mathrm{in}}})$ \label{alg:iselect}
\EndFunction
\Function{BootstrapHalfWidth}{$\ell, \tau$} \Comment{$\ell$: list of losses}
    \For{$b = 1$ to $N_{\mathrm{boot}}$}
        \State $\bar{\ell}_b$ $\gets$ mean of $|\ell|$ losses drawn from $\ell$ with replacement \label{alg:bdraw}
    \EndFor
    \State \Return half the distance between the $\tau/2$ and $(1-\tau/2)$ quantiles of $(\bar{\ell}_1, \cdots, \bar{\ell}_{N_{\mathrm{boot}}})$
    \label{alg:bret}
\EndFunction
\State Split every $\mathcal{D}_g$ into $K$ folds $S_{g,1}, \cdots, S_{g,K}$ \label{alg:fsplit}
\For{$g = 1$ to $G$}
    \For{$k = 1$ to $K$} \Comment{outer level}
        \State $\mathcal{D}^{\mathrm{in}}_{g,k} \gets \mathcal{D}_g \setminus S_{g,k}$ \label{alg:fin}
        \State $a_{g,k} \gets \Call{SelectPredictor}{\mathcal{D}^{\mathrm{in}}_{g,k}}$ \label{alg:fselect}
        \State Train $a_{g,k}$ on $\mathcal{D}^{\mathrm{in}}_{g,k}$ to obtain $\hat{f}_{g,k}$ \label{alg:ffit}
        \State $\ell_{g,k} \gets \{\Loss(\hat{f}_{g,k}; x, y) : (x, y) \in S_{g,k}\}$ \label{alg:fscore}
    \EndFor
    \State $\Est{g} \gets$ mean of $(\ell_{g,1}, \cdots, \ell_{g,K})$ \label{alg:fest}
    \State $c_g \gets \Call{BootstrapHalfWidth}{(\ell_{g,1}, \cdots, \ell_{g,K}), \beta/G}$ \label{alg:fmargin}
    \State $\RefPt{g} \gets \max\{\Est{g} - c_g, 0\}$ \label{alg:fref}
    % \label{alg:fref}  (duplicate label, removed)
\EndFor
\State \Return $\{\RefPt{g}\}_{g=1}^G$
\end{algorithmic}
\end{algorithm}

Specifically, for each group we define a set $\mathcal{A}$ of candidate model architectures, e.g., linear models and small neural networks (Appendix~\ref{app:floors}). We then split the training patients of the group into $K$ folds, as shown in line~\ref{alg:fsplit}. In each of the $K$ iterations, one fold is held out for the final evaluation, and the remaining $K-1$ folds are used to choose the architecture and train the model, as shown in line~\ref{alg:fin}. To choose the architecture, we perform a second cross-validation on the remaining folds, with $K_{\mathrm{in}}$ inner folds. Every architecture in $\mathcal{A}$ is trained on all inner folds but one and evaluated on the remaining one. The architecture with the lowest average loss is selected, as shown in the function \textsc{SelectPredictor} (lines~\ref{alg:isplit}--\ref{alg:iselect}) and called in line~\ref{alg:fselect}. Once the architecture is selected, we train the model on the $K-1$ remaining folds and compute its loss on the held-out fold, as shown in lines~\ref{alg:ffit}--\ref{alg:fscore}. We need two layers of cross-validation because choosing an architecture and evaluating it on the same patients would make the estimate too optimistic \citep{varma2006bias,tibshirani2009bias}, which is also shown in Proposition~\ref{prop:estimate}. The estimate $\Est{g}$ is the average of the held-out losses over the $K$ iterations, as shown in line~\ref{alg:fest}. Finally, since $\Est{g}$ is an average over a finite number of patients, it is noisy. A reference that is too high would hide part of the excess loss of the group and lead to underweighting it. To guard against this, we lower $\Est{g}$ by a margin $c_g$ that reflects its uncertainty, which is the half-width of a bootstrap confidence interval for $\Est{g}$ \citep{efron1993introduction}. The reference then becomes
\begin{equation}
\RefPt{g}=\max\{\Est{g}-c_g,0\},
\label{eq:refpt}
\end{equation}
since a loss cannot be negative, as shown in lines~\ref{alg:fmargin}--\ref{alg:fref}.
The bootstrap resamples the held-out losses of the group with replacement and computes the average of each resample, as shown in the function \textsc{BootstrapHalfWidth} (lines~\ref{alg:bdraw}--\ref{alg:bret}). The interval contains the middle $1-\beta/G$ fraction of these averages, where $\beta$ is the error we allow over all groups together, so each of the $G$ intervals may miss the true value with probability $\beta/G$ \citep{dunn1961multiple}. The margin is larger for groups with fewer patients, since their averages are noisier. This confidence level is only approximate, and we use the margin as a heuristic correction and not as a confidence bound for $\Ach{g}$. Appendix~\ref{app:floors} gives additional details, shows that the errors of the references affect the solution only through their spread (Proposition~\ref{prop:spread}), and describes the simplified procedure that we use on EMBED, which has no margin and is not covered by Proposition~\ref{prop:estimate}.

\subsection{Training Algorithm}
\label{sec:training}
\begin{algorithm}[t]\small
\caption{Excess-loss group DRO with latent anchors}
\label{alg:train}
\begin{algorithmic}[1]
\Require Training data $\mathcal{D}_g$ of every group $g$, references $\RefPt{g}$, number of iterations $T$, learning rate $\eta_\theta$, step size $\gamma$, smoothing $\rho$, update interval $T_\lambda$, weights $\alpha_{\mathrm{align}}, \alpha_{\mathrm{sep}}$, number of anchor samples $M$
\State Initialize $\theta = (\{\phi_g\}, \psi, \{\mu_c\})$ randomly \label{alg:init}
\State $\lambda_g \gets 1/G$ and $\bar{\Loss}_g \gets \RefPt{g}$ $\forall g$ \label{alg:initw}
\For{$t = 1$ to $T$}
    \State Sample a minibatch $B_g$ of patients from $\mathcal{D}_g$ $\forall g$ \label{alg:batch}
    \State $\hat{\Loss}_g \gets$ loss of $\psi \circ \phi_g$ on $B_g$ $\forall g$ \label{alg:task}
      \State $\hat{\nu}_g^c \gets$ diagonal Gaussian fitted to $\{\phi_g(x) : (x, y) \in B_g, y = c\}$ $\forall g$, $c \in \mathcal{C}_g$ \label{alg:nu} \label{alg:nu}
    \State $\hat{\Loss}^{\mathrm{align}}_g \gets \frac{1}{|\mathcal{C}_g|}\sum_{c \in \mathcal{C}_g}\Wtwo^2(\hat{\nu}_g^c, \mu_c)$ $\forall g$ \label{alg:align}
    \State $\hat{\Loss}^{\mathrm{sep}} \gets \frac{1}{CM}\sum_{c}\sum_{r=1}^{M} \Loss(\psi; z_{c,r}, c)$, with $z_{c,r} \sim \mu_c$ \label{alg:sep}
    \State $Q \gets \sum_g \lambda_g \big(\hat{\Loss}_g + \alpha_{\mathrm{align}}\hat{\Loss}^{\mathrm{align}}_g\big)$ \label{alg:objective}
    \State $d_\phi \gets \nabla_\phi Q$, $d_\psi \gets \nabla_\psi \sum_g \lambda_g \hat{\Loss}_g$, $d_\mu \gets \nabla_\mu \big(Q + \alpha_{\mathrm{sep}}\hat{\Loss}^{\mathrm{sep}}\big)$ \label{alg:grad}
    \State $\phi \gets \phi - \eta_\theta d_\phi$, $\psi \gets \psi - \eta_\theta d_\psi$, $\mu \gets \mu - \eta_\theta d_\mu$ \Comment{update encoders, head, anchors} \label{alg:descent}
    \State $\bar{\Loss}_g \gets \rho\,\bar{\Loss}_g + (1-\rho)\big(\hat{\Loss}_g + \alpha_{\mathrm{align}}\hat{\Loss}^{\mathrm{align}}_g\big)$ $\forall g$ \label{alg:average}
    \If{$t \bmod T_\lambda = 0$} \Comment{update the weights}
        \State $\lambda'_g$ $\gets \lambda_g \exp\big\{\gamma(\bar{\Loss}_g - \RefPt{g})\big\}$ $\forall g$ \label{alg:mirror}
        \State $\lambda_g \gets$ $\lambda'_g / \sum_{g'=1}^G \lambda'_{g'}$ $\forall g$ \label{alg:normalize}
    \EndIf
\EndFor
\State \Return model parameters $\theta$
\label{alg:return}
\end{algorithmic}
\end{algorithm}
We propose Algorithm~\ref{alg:train} to train the model on the final problem~\eqref{eq:objective}, with $\RefPt{g}$ in place of $\Ach{g}$. The algorithm alternates between updating the model (the encoders, the head and the class anchors) and updating the weights of the groups. We first initialize the parameters of the model $\theta$ randomly and the weights $\lambda$ of the groups uniformly, as shown in lines~\ref{alg:init}--\ref{alg:initw}. Then, at each iteration, we sample a minibatch with patients from every group, as shown in line~\ref{alg:batch}, and estimate on it the three losses of the final problem \eqref{eq:objective}. The first is the task loss $\hat{\Loss}_g$ of each group, which estimates \eqref{eq:grouploss} (line~\ref{alg:task}). The second is the alignment loss $\hat{\Loss}^{\mathrm{align}}_g$ \eqref{eq:align}, where, for each group and each class, the diagonal Gaussian distribution $\hat{\nu}_g^c$ is estimated from the mean and the variance of the embeddings of the corresponding patients in the minibatch and compared with the current anchor $\mu_c$ (lines~\ref{alg:nu}--\ref{alg:align}). Since a class with fewer than two patients in the minibatch has no variance estimate, the average over the classes runs only over the set $\mathcal{C}_g$ of the classes with at least two patients of group $g$, and the loss is zero if there is none. The details of this estimator are in Appendix~\ref{app:anchor-impl}. The third is the separation loss $\hat{\Loss}^{\mathrm{sep}}$ \eqref{eq:sep}, which is computed on $M$ samples from each of the current anchors (line~\ref{alg:sep}). The task and alignment losses of each group are multiplied by its weight $\lambda_g$ and summed in $Q$ (line~\ref{alg:objective}). Then, each block of parameters is updated with its own gradient (lines~\ref{alg:grad}--\ref{alg:descent}). The encoders are updated with the gradient of $Q$, the anchors with the gradient of $Q$ plus the separation loss, and the head with the gradient of the weighted task loss alone. The shared head is not updated by the separation loss{, since it should be trained from patients and not from samples drawn from the anchors, which do not correspond to real patients. Hence, these updates are not gradient descent on \eqref{eq:objective}. In the experiments, we use Adam (AdamW on EMBED) instead of the plain gradient steps of line~\ref{alg:descent} (Appendix~\ref{app:training}). Every $T_\lambda$ iterations, we update the weights of the groups. Since the loss of a group is estimated on a small minibatch, which makes it noisy, we keep a running average $\bar{\Loss}_g$ of the loss of each group, including its alignment term, so that the updates are smoother (line~\ref{alg:average}). It is an exponential moving average with smoothing parameter $\rho$, and it is initialized at the references, so that every group starts with zero excess loss. Then, the weights are updated with a mirror ascent step {on the excess loss $\bar{\Loss}_g - \RefPt{g}$ (lines~\ref{alg:mirror}--\ref{alg:normalize}).
\begin{table*}[t]
\caption{Comparison with the baselines on the original groups (left) and with disjoint feature spaces (right). WG: worst group; exc.: excess loss with respect to the references; $\Delta$: largest minus smallest value across groups. Mean over $10$ seeds ($5$ folds in Fed-Heart). Params: parameters at inference, in thousands. Best in bold. $^*$ ($^\dagger$): ours is significantly better (worse) than this method on that metric (paired $t$-test over the 10 seeds, $p<0.05$, two-sided, without correction for multiple comparisons; the folds of Fed-Heart are pooled within each seed).}
\label{tab:main}
\centering
\footnotesize
\setlength{\tabcolsep}{2.2pt}
\resizebox{\textwidth}{!}{%
\begin{tabular}[t]{@{}c@{\hspace{2.5em}}c@{}}
\textbf{Original groups} & \textbf{Disjoint feature spaces}\\[2pt]
\begin{tabular}[t]{@{}lccccccc@{}}
\toprule
Method & \shortstack{WG\\acc$\uparrow$} & \shortstack{$\Delta$\\acc$\downarrow$} & \shortstack{WG\\loss$\downarrow$} & \shortstack{$\Delta$\\loss$\downarrow$} & \shortstack{WG\\exc.$\downarrow$} & \shortstack{$\Delta$\\exc.$\downarrow$} & \shortstack{Params\\(K)$\downarrow$} \\
\midrule
\multicolumn{8}{@{}l}{\textit{NHANES}} \\
ERM (common) & 65.0 & 5.9 & 0.551 & 0.077 & 0.295 & 0.063 & \textbf{11.4} \\
GroupDRO (common) & \textbf{66.5} & \textbf{5.8} & \textbf{0.523} & \textbf{0.075} & \textbf{0.262} & \textbf{0.057} & \textbf{11.4} \\
Dedicated & 60.8 & 11.0 & 0.596 & 0.132 & 0.317 & 0.095 & 35.1 \\
Reweigh & 62.1$^*$ & 10.0 & 0.606$^*$ & 0.130 & 0.327$^*$ & 0.093 & 52.5 \\
Flex-MoE & 62.0$^*$ & 9.3 & 0.597$^*$ & 0.125 & 0.318$^*$ & 0.088 & 536.4 \\
REMIND & 65.2 & 11.1 & 0.561 & 0.139 & 0.282 & 0.103 & 53.2 \\
\textbf{Ours} & 65.7 & 8.5 & 0.548 & 0.109 & 0.280 & 0.073 & 30.8 \\
\midrule
\multicolumn{8}{@{}l}{\textit{Fed-Heart}} \\
ERM (common) & 64.4 & 12.9 & 0.628 & 0.149 & 0.518 & 0.415 & \textbf{11.6} \\
GroupDRO (common) & 65.9 & \textbf{11.4} & 0.619 & \textbf{0.110} & 0.452 & 0.319 & \textbf{11.6} \\
Dedicated & 74.8 & 18.7 & 0.572 & 0.239 & 0.257 & 0.156 & 45.4 \\
Reweigh & 75.7$^\dagger$ & 17.8 & 0.562 & 0.277 & 0.210 & 0.141 & 53.5 \\
Flex-MoE & \textbf{76.1}$^\dagger$ & 17.2 & 0.591$^*$ & 0.118 & 0.415$^*$ & 0.324 & 537.2 \\
REMIND & 75.5$^\dagger$ & 18.0 & 0.570 & 0.256 & 0.238$^*$ & 0.169 & 54.5 \\
\textbf{Ours} & 74.4 & 19.1 & \textbf{0.561} & 0.280 & \textbf{0.206} & \textbf{0.103} & 38.9 \\
\midrule
\multicolumn{8}{@{}l}{\textit{EMBED}} \\
Dedicated & 41.4 & 36.7 & 1.287 & 0.777 & 0.390 & 0.488 & \textbf{275.7} \\
Reweigh & \textbf{64.5} & \textbf{12.5} & 3.685$^*$ & 3.157 & 2.982$^*$ & 3.071 & 1288.6 \\
Flex-MoE & 61.4 & 16.0 & 2.442$^*$ & 1.916 & 1.740$^*$ & 1.814 & 927.5 \\
REMIND & 64.2 & 12.7 & 3.042$^*$ & 2.517 & 2.339$^*$ & 2.433 & 1337.7 \\
\textbf{Ours} & 62.2 & 14.0 & \textbf{1.052} & \textbf{0.534} & \textbf{0.240} & \textbf{0.318} & \textbf{275.7} \\
\bottomrule
\end{tabular}
&
\begin{tabular}[t]{@{}lccccccc@{}}
\toprule
Method & \shortstack{WG\\acc$\uparrow$} & \shortstack{$\Delta$\\acc$\downarrow$} & \shortstack{WG\\loss$\downarrow$} & \shortstack{$\Delta$\\loss$\downarrow$} & \shortstack{WG\\exc.$\downarrow$} & \shortstack{$\Delta$\\exc.$\downarrow$} & \shortstack{Params\\(K)$\downarrow$} \\
\midrule
\multicolumn{8}{@{}l}{\textit{NHANES (designed)}} \\
Dedicated & 56.0 & 16.0 & 0.679 & 0.202 & 0.356 & 0.130 & 33.6 \\
Reweigh & 56.3$^*$ & 14.2 & 0.696$^*$ & 0.213 & 0.373$^*$ & 0.141 & 52.5 \\
Flex-MoE & 59.8$^*$ & 10.8 & 0.687$^*$ & 0.183 & 0.364$^*$ & 0.111 & 536.4 \\
REMIND & 62.8 & 12.3 & 0.622$^*$ & 0.186 & 0.299$^*$ & 0.114 & 53.2 \\
\textbf{Ours} & \textbf{64.5} & \textbf{7.4} & \textbf{0.590} & \textbf{0.123} & \textbf{0.278} & \textbf{0.050} & \textbf{29.3} \\
\midrule
\multicolumn{8}{@{}l}{\textit{Fed-Heart}} \\
Dedicated & \textbf{64.0} & \textbf{28.8} & \textbf{0.626} & 0.398 & 0.201 & 0.146 & 44.0 \\
Reweigh & 63.3 & 29.4 & 0.628$^\dagger$ & 0.423 & \textbf{0.178}$^\dagger$ & \textbf{0.120} & 52.5 \\
Flex-MoE & 63.2 & 29.3 & 0.633 & 0.422 & 0.183$^\dagger$ & 0.122 & 535.7 \\
REMIND & 63.6 & 29.1 & 0.627$^\dagger$ & 0.416 & 0.185$^\dagger$ & 0.129 & 53.6 \\
\textbf{Ours} & 63.4 & 29.4 & 0.635 & \textbf{0.389} & 0.219 & 0.154 & \textbf{37.5} \\
\midrule
\multicolumn{8}{@{}l}{\textit{EMBED}} \\
Dedicated & 53.6 & 19.6 & 1.153 & 0.559 & 0.323 & 0.490 & \textbf{230.4} \\
Reweigh & 58.8 & 14.4 & 4.984$^*$ & 4.392 & 4.152$^*$ & 4.325 & 1288.1 \\
Flex-MoE & \textbf{60.8} & \textbf{12.1} & 2.311$^*$ & 1.661 & 1.498$^*$ & 1.593 & 927.5 \\
REMIND & 58.5 & 14.8 & 4.297$^*$ & 3.708 & 3.465$^*$ & 3.640 & 1320.8 \\
\textbf{Ours} & 58.0 & 15.3 & \textbf{1.051} & \textbf{0.442} & \textbf{0.219} & \textbf{0.375} & \textbf{230.4} \\
\bottomrule
\end{tabular}
\end{tabular}}
\end{table*}

\providecommand{\newtxt}[1]{\textcolor{blue}{#1}}
\providecommand{\deltxt}[1]{\textcolor{blue}{\sout{#1}}}
\renewcommand{\newtxt}[1]{\textcolor{blue}{#1}}
\renewcommand{\deltxt}[1]{\textcolor{blue}{\sout{#1}}}

\section{Experiments}
In this section, we introduce the datasets that we use for our experiments, the baselines that we compare against, and the protocol that all the methods follow. The loss $\mathcal{L}$ is the cross-entropy in all the experiments. All the implementation details are in Appendix~\ref{app:details}, and the ablation and synthetic experiments are in Appendices~\ref{app:ablations} and~\ref{app:synthetic}.

\subsection{Datasets and Setting Designs}
\label{sec:datasets}
We run experiments on three public medical datasets, NHANES, Fed-Heart and EMBED. In all of them, the groups are a fact of how the data were collected and not a split that we impose. For each dataset, we describe the task, the rate of the classes and how the groups are defined. Table~\ref{tab:datasets} in Appendix~\ref{app:datasets} gives, for every group, its number of patients and its features, and Appendix~\ref{app:training} gives the splits.
\textbf{NHANES} \citep{nhanes_2017_2020,nhanes_2021_2023} is the National Health and Nutrition Examination Survey of the CDC. It contains $17{,}005$ US adults. The task is to predict cardiovascular disease, which $10.5\%$ of the participants have. The groups are defined by how much of the assessment a participant completed.
\textbf{Fed-Heart} \citep{duterrail2022flamby,uci_heart_disease_1989} contains $920$ patients from four cardiology hospitals. The task is to predict whether a patient has heart disease, which $55.3$\% of the patients have. The groups are the hospitals, which recorded different subsets of the same clinical features.
\textbf{EMBED} \citep{jeong2023emory} contains $128{,}680$ breast-exam records from $22{,}997$ patients. The task is to predict the BI-RADS breast density, which has four ordered classes, with proportions $10.3\%$, $42.1\%$, $42.3\%$ and $5.4\%$ from A to D, which we treat as a four-class classification problem. The groups are defined by which of the four imaging views an exam has, and the features of a view are the embedding of a frozen ViT-Base (Appendix~\ref{app:models}). The group sizes range from $65$ to $73{,}040$ records, and no view is recorded by every group, so the common-feature baselines of Section~\ref{sec:baselines} are not available on EMBED.

\textbf{Designs with disjoint feature spaces.} For each dataset, we also design an experiment in which the groups have fully disjoint feature spaces, by assigning to each group its own subset of the features (Table~\ref{tab:datasets}). The patients and the groups are those of the original setting, and only the features that each group may use change; the blocks of features are fixed before any training. These designs test the two situations that our framework targets: groups that can only share information through the latent space and the head, since no test is shared, and groups whose optimal losses differ. In NHANES, the three blocks are chosen so that the optimal losses of the groups differ more than in the original groups: the references are $0.323$, $0.251$ and $0.285$, a spread of $0.07$ against $0.04$ in the original groups (Appendix~\ref{app:datasets}, Table~\ref{tab:refvalues}). In EMBED, two groups are dropped, since four views cannot form six disjoint groups, which is why the number of parameters at inference differs between the two settings in Table~\ref{tab:main}.

\subsection{Baselines}
\label{sec:baselines}

We compare against standard approaches for these settings. First, we train a dedicated model for each group on its own features. Second, we train a model on the common feature space, discarding all the other information, using either ERM or group DRO as the optimization technique. The common feature models are only available when the groups share at least one feature. Finally, we compare against state-of-the-art methods that account for the heterogeneous feature spaces setting: Flex-MoE \citep{yun2024flexmoe}, REMIND \citep{wu2026remind}, and Reweigh (the inverse-frequency re-weighting baseline of \citet{wu2026remind}). All the baselines use the same splits, seeds and training protocol as our framework (Appendix~\ref{app:baselines}).

\providecommand{\newtxt}[1]{\textcolor{blue}{#1}}
\providecommand{\deltxt}[1]{\textcolor{blue}{\sout{#1}}}
\renewcommand{\newtxt}[1]{\textcolor{blue}{#1}}
\renewcommand{\deltxt}[1]{\textcolor{blue}{\sout{#1}}}

\section{Results}
\label{sec:results}
\subsection{Comparison with the Baselines}
\label{sec:overall}
Table~\ref{tab:main} (left) compares our framework with the baselines when the groups keep their original features. Among the methods that account for heterogeneous feature spaces (Reweigh, Flex-MoE and REMIND), ours has the lowest worst-group loss and the lowest excess loss on every dataset, with a significantly lower excess loss in seven of the nine comparisons (e.g., in EMBED, its worst-group loss is $1.052$ against $2.442$ to $3.685$). Furthermore, our framework does this while using $1.4\times$ to $17.4\times$ fewer parameters than these methods, and up to $18\times$ fewer in the designs with disjoint feature spaces. The only baseline with a lower loss is group DRO on the common features alone in NHANES ($0.523$ against $0.548$), which suggests that, in NHANES, the common features already carry most of the information about the outcome; consistently, the references of its three groups are close to each other (Table~\ref{tab:refvalues}). When it comes to the worst-group accuracy, our framework is comparable to the baselines, although not always the best. It is at most $2.3$ points below the best baseline ($74.4\%$ against $76.1\%$ in Fed-Heart, and $62.2\%$ against $64.5\%$ in EMBED). It is, however, the only method that, in every dataset, has both a low worst-group loss and a worst-group accuracy close to the best: in EMBED, the baselines with the highest accuracy have a worst-group loss above $3$, and the dedicated models have a low loss but an accuracy of $41.4\%$.

\subsection{Disjoint Feature Spaces}
\label{sec:disjoint}
Table~\ref{tab:main} (right) shows the settings with disjoint feature spaces. We designed the NHANES setting so that the optimal losses of the groups differ more than in the original groups (Section~\ref{sec:datasets}); there, ours outperforms the other baselines in all the metrics, with a worst-group excess loss of $0.278$ against $0.299$ to $0.373$. In a setting like EMBED, where some groups have too few records to be trained on their own, we also outperform the other baselines in the losses; in both settings, the differences in the losses are significant against Reweigh, Flex-MoE and REMIND, and not against the dedicated models. In Fed-Heart, in contrast, the hospitals are small, so there is little to transfer: a dedicated model is as good as sharing, and ours has a higher excess loss than the baselines (0.219 against 0.178 to 0.201). A likely reason is that the references of the smallest hospitals are unreliable (Table~\ref{tab:refvalues}: the bootstrap margin lowers the reference of Switzerland, with $123$ patients, from $0.237$ to $0.027$), and that the anchors raise the excess loss in this setting (Appendix~\ref{app:ablations}). The ablation and synthetic experiments (Appendices~\ref{app:ablations} and~\ref{app:synthetic}) show that the excess loss helps when the optimal losses differ and are estimated reliably, whereas the benefit of the anchors on the losses is mixed and depends on whether some groups cannot train a good model on their own.

\subsection{Anchors and the Latent Space}
\label{sec:latent}
\begin{table}[t]
\caption{Latent space of the test patients without and with the anchors. $W_2$: 2-Wasserstein distance between the embeddings of two groups (two classes), divided by the standard deviation of all the test embeddings and averaged over all pairs, so that the distance does not depend on the scale of the latent space, which the anchors shrink (Appendix~\ref{app:latent_space}). Nearest-anchor acc.: percentage of patients closest to the anchor of their class (class centroid without anchors). Best in bold.}
\label{tab:latent}
\centering
\footnotesize
\setlength{\tabcolsep}{4pt}
\resizebox{\linewidth}{!}{%
\begin{tabular}{@{}llccc@{}}
\toprule
 & & \multicolumn{2}{c}{$W_2$} & \\
\cmidrule(lr){3-4}
Dataset & Model & \shortstack{groups\\$\downarrow$} & \shortstack{classes\\$\uparrow$} & \shortstack{Nearest-\\anchor acc.$\uparrow$} \\
\midrule
\multirow{2}{*}{NHANES} & No anchors & 0.53 & 0.52 & \textbf{68.4} \\
 & Anchors & \textbf{0.03} & \textbf{0.53} & \textbf{68.4} \\
\midrule
\multirow{2}{*}{Fed-Heart} & No anchors & 1.08 & 0.29 & \textbf{79.5} \\
 & Anchors & \textbf{0.33} & \textbf{0.44} & 77.3 \\
\midrule
\multirow{2}{*}{EMBED} & No anchors & 0.78 & \textbf{0.66} & 67.0 \\
 & Anchors & \textbf{0.27} & 0.57 & \textbf{76.0} \\
\bottomrule
\end{tabular}}
\end{table}

Table~\ref{tab:latent} compares the latent space of the test patients with and without the alignment and separation losses, i.e., our framework against the Excess DRO variant of Appendix~\ref{app:ablations}, which has the same architecture and objective but $\alpha_{\mathrm{align}} = \alpha_{\mathrm{sep}} = 0$. Without them, nothing prevents each encoder from mapping its group to its own region (Proposition~\ref{prop:separate}). We measure this with the $W_2$ distance between groups, which shows whether each group lies in its own region. Across all settings we observe that when anchors are used, the distance between groups decreases by $3\times$ to $18\times$. A small distance between groups could also come from a collapse of all the representations, which the separation loss is designed to prevent. For this, we also measure the distance between classes, and observe that it increases in NHANES (0.52 to 0.53) and Fed-Heart (0.29 to 0.44) and decreases in EMBED (0.66 to 0.57), so the representations do not collapse. Finally, we want to see that the latent space is properly organized according to the class. For this we measure the percentage of patients whose embedding is closest to their own class, which we denote as nearest-anchor accuracy. We observe that it improves in a setting where the features are quite heterogeneous, as in EMBED, remains the same in NHANES, and drops slightly in the tabular case with few patients, Fed-Heart. Overall, with the anchors the groups overlap and the classes are organized around their anchors, although in EMBED the two largest groups still spread along different directions (Appendix~\ref{app:latent_space}). We do not evaluate the uses of such a latent space, such as transferring models between groups or explaining predictions by their position relative to the anchors; Appendix~\ref{app:latent_space} discusses them.

\section{Conclusion}
We studied group-robust learning across heterogeneous feature spaces. We proposed to encode each group into a common latent space organized around class anchors, and to predict from it with a single shared head. We trained the model to minimize the worst-group excess loss, with the optimal losses estimated by nested cross-validation. We conducted experiments on three medical datasets, together with ablation studies and synthetic experiments, and show that our method outperforms state-of-the-art methods, especially when groups differ in their optimal losses and when some of them cannot train a good model on their own but benefit from collaborating.

\newpage
% ---------------------------------------------------------------------
\bibliographystyle{plainnat}
\bibliography{refs}

% ---------------------------------------------------------------------
\clearpage
\onecolumn
\appendix
\section{Implementation Details}
\label{app:details}

\subsection{Datasets}
\label{app:datasets}

\begin{center}
\begin{minipage}{\textwidth}
\centering
\captionof{table}{Datasets, groups and features. For each group, we show its number of samples, and its features in the original data and in the design with disjoint feature spaces. The number of samples of the groups of EMBED is the number of records. CC and MLO denote the craniocaudal and the mediolateral oblique views.}
\label{tab:datasets}
\footnotesize
\setlength{\tabcolsep}{3.5pt}
\begin{tabular}{@{}p{2.3cm}p{2.5cm}p{5.3cm}p{5.0cm}@{}}
\toprule
Dataset & Group (samples) & Original features & Disjoint features \\
\midrule
\multirow{3}{*}{\shortstack[l]{NHANES\\$17{,}005$ adults}} & G0 ($2{,}666$) & Questionnaire (10) & Race, education, income, smoking \\
 & G1 ($2{,}648$) & Questionnaire, body measures (13) & Age, gender, body measures \\
 & G2 ($11{,}691$) & Questionnaire, body measures, blood pressure, blood tests (20) & Blood tests, blood pressure \\
\midrule
\multirow{4}{*}{\shortstack[l]{Fed-Heart\\$920$ patients}} & Cleveland ($303$) & 10 of the 13 clinical features & Demographics, resting BP \\
 & Hungarian ($294$) & 8 of the 13 clinical features & Exercise test \\
 & Switzerland ($123$) & 8 of the 13 clinical features & Chest-pain type \\
 & VA ($200$) & 9 of the 13 clinical features & Blood tests, resting ECG \\
\midrule
\multirow{6}{*}{\shortstack[l]{EMBED\\$22{,}997$ patients\\$128{,}680$ records}} & g1 ($1{,}158$) & FFDM CC & FFDM CC \\
 & g2 ($187$) & C-View CC, FFDM CC & C-View CC \\
 & g3 ($543$) & FFDM MLO & FFDM MLO \\
 & g4 ($73{,}040$) & FFDM CC, FFDM MLO & (dropped) \\
 & g5 ($65$) & C-View CC, FFDM CC, FFDM MLO & (dropped) \\
 & g6 ($53{,}687$) & C-View CC, C-View MLO, FFDM CC, FFDM MLO & C-View MLO \\
\bottomrule
\end{tabular}
\end{minipage}
\end{center}

Table~\ref{tab:datasets} gives the groups of each dataset, with their number of samples and their features in the original data and in the design with disjoint feature spaces.

\textbf{NHANES.} The raw files of the two CDC survey cycles are public. After the preprocessing, each participant has 20 features: age, gender, race (five one-hot features), education, income-to-poverty ratio, smoking status, BMI, weight, height, the mean systolic and diastolic blood pressure, HbA1c, HDL, total cholesterol, triglycerides and LDL.

\textbf{Fed-Heart.} The four raw files of the UCI heart disease archive are public. We keep 13 features: age, sex, resting blood pressure, serum cholesterol, fasting blood sugar, maximum heart rate, exercise-induced angina, ST depression, the type of chest pain (three one-hot features) and the resting ECG (two one-hot features), and we impute the missing values with the median. In the design with disjoint feature spaces, the four blocks of features are fixed before running any experiment. All the reported results use the full training set of every hospital.

\textbf{EMBED.} The images are distributed under a research use agreement that does not allow sharing them, or any data derived from them such as the embeddings. The features of a view are the 768-dimensional embedding of a frozen ViT-Base. In the design with disjoint feature spaces, g4 and g5 are dropped, since four views cannot form six disjoint groups.

\textbf{Designed NHANES.} In the design with disjoint feature spaces, the participants and the groups are those of the original setting, and only the features that each group may use change. The 20 features are assigned to the groups in three disjoint blocks, fixed before any training so that the achievable losses of the groups differ more than in the original groups. G0 uses race (five one-hot features), education, the income-to-poverty ratio and smoking status (8 features), and deliberately not age or gender. G1 uses age, gender, BMI, weight and height (5 features). G2 uses HbA1c, HDL, systolic and diastolic blood pressure, total cholesterol, triglycerides and LDL (7 features). No feature is shared, so the common-features models are not available. The references, estimated once on the training rows as in Appendix~\ref{app:floors}, are $0.323$, $0.251$ and $0.285$ for G0, G1 and G2, a spread of $0.07$ against $0.04$ in the original groups.

\subsection{Models}
\label{app:models}

\textbf{NHANES and Fed-Heart.} Each group has its own encoder, an MLP with layer normalization whose input has the width of the features of the group, and whose output is a latent vector of dimension $64$ (two hidden layers of width $64$, ReLU, dropout $0.1$). The head, shared by all groups, is an MLP with one hidden layer of $32$ units (ReLU, dropout $0.3$) that maps the latent space to the classes.

\textbf{EMBED.} The $768$-dimensional embedding of each view is projected to $64$ dimensions by a linear layer shared by all groups. Each group has an MLP (hidden width $64$, GELU) that takes the concatenation of the projections of its views and outputs a latent vector of dimension $64$. The head is a linear layer from the latent space to the four classes.

In all the datasets, there is one anchor per class, a Gaussian with diagonal covariance in the latent space, which adds $2 \times 64 \times C$ parameters. The anchors are only used for training, so the parameters at inference are those of the encoders and the head: $30.8$K, $38.9$K and $275.7$K in NHANES, Fed-Heart and EMBED, and $29.3$K, $37.5$K and $230.4$K in the designs with disjoint feature spaces.

\subsection{Training}
\label{app:training}

We train our framework and the baselines with the same protocol, which was fixed before running the experiments. Table~\ref{tab:hyper} gives the hyperparameters. In NHANES and Fed-Heart, every batch keeps the proportions of the groups in the training set, and in EMBED every batch has the same number of records from each group, where ERM weights the groups by their proportions. The group weights start uniform and are updated every $T_\lambda$ steps with the exponentiated update of Algorithm~\ref{alg:train}, using a running average of the loss of the training batches, including the alignment term. The excess loss is signed, i.e., we do not clip it at zero. The separation loss updates the anchors only, and the head is trained by the task loss alone. The running average starts at the reference, which is zero for the GroupDRO variants. The gradient norm is clipped at $1.0$ in NHANES and Fed-Heart. In NHANES, every method uses inverse-frequency class weights, and in Fed-Heart none does. In Fed-Heart, the imputation and the references of each fold only use its training patients.

We train for a fixed number of epochs without early stopping, we record the metrics after every epoch, and we report the test metrics of epoch $10$ for all the methods, so no quantity of the validation or the test sets is used to choose the epoch. We repeat every experiment with $10$ seeds, and with $5$ folds in Fed-Heart. The tabular experiments ran on an Apple M3 Pro laptop CPU, and the EMBED experiments on one NVIDIA A10G GPU on an AWS instance.

\begin{table*}[t]
\caption{Hyperparameters of our framework. The baselines use the same optimizer settings, batches, epochs and class weights.}
\label{tab:hyper}
\centering
\footnotesize
\setlength{\tabcolsep}{6pt}
\begin{tabular}{@{}llll@{}}
\toprule
 & NHANES & Fed-Heart\hspace{1.2em} & EMBED \\
\midrule
Optimizer & Adam & Adam & AdamW \\
Learning rate & \multicolumn{2}{c}{$10^{-3}$ to $10^{-5}$, cosine} & $5\cdot10^{-5}$, $\times0.1$ every 5 epochs \\
Weight decay & $10^{-4}$ & $10^{-4}$ & $5\cdot10^{-5}$ \\
Gradient clipping & $1.0$ & $1.0$ & none \\
Batch size & 128 & 64 & 32 per group \\
Epochs (reported) & 30 (10) & 30 (10) & 20 (10) \\
Class weights & inverse-frequency & none & none \\
$\alpha_{\mathrm{align}} = \alpha_{\mathrm{sep}}$ & 0.1 & 0.1 & 1.0 \\
$\rho$, $T_\lambda$ & \multicolumn{2}{c}{$0.9$, $\min(50, \text{steps per epoch})$} & $0.9$, $50$ \\
Step size $\gamma$ (grid) & \multicolumn{2}{c}{$\{0.1, 0.5, 2, 10\}$} & $\{0.5, 2\}$ \\
Seeds & 10 & 10 (5 folds) & 10 \\
\bottomrule
\end{tabular}
\end{table*}

\subsection{Hyperparameters}
\label{app:hyper}

The step size $\gamma$ of the weight update is the only hyperparameter that we tune. For each method and dataset, we run all the values of a grid fixed in advance (Table~\ref{tab:hyper}) with the $10$ seeds, and we choose the value with the lowest mean worst-group loss on the validation set at epoch $10$. The architecture, the optimizer, the batch size and the weights $\alpha_{\mathrm{align}}$ and $\alpha_{\mathrm{sep}}$ are fixed and the same for all the variants. The selected step size of our framework is $2$ in NHANES and EMBED, $0.1$ in Fed-Heart, and, in the designs with disjoint feature spaces, $0.1$ in NHANES, $0.1$ in Fed-Heart and $2$ in EMBED. In Fed-Heart, the validation set of each fold is $15\%$ of its training patients, and the step size is chosen on the mean of the validation worst-group loss over seeds and folds. No quantity of the test set is used in any decision.

\subsection{Baselines}
\label{app:baselines}

\textbf{Dedicated models and common features.} A dedicated model is a network trained on each group, with its own head, that uses all the features of the group. The common-features models have a single encoder and head, trained on the features that all groups have, with either ERM or group DRO.

\textbf{Soft MoE.} Reweigh and REMIND share the same backbone \citep{puigcerver2023sparse}, a mixture of experts with soft routing: each expert receives a weighted combination of the inputs, and the outputs of the experts are combined with learned weights, so the routing is differentiable. Every test (or view) has its own encoder, and the tests that a group does not have are replaced by a learnable embedding.

\textbf{Reweigh} trains this backbone with the loss of each group weighted by the inverse of its frequency; on EMBED, where the batches are balanced, the weights are uniform. The REMIND paper reports this baseline without implementation details, so we use the standard inverse-frequency re-weighting. \textbf{Flex-MoE} \citep{yun2024flexmoe} is a sparse mixture of experts: a bank gives a learnable embedding to the missing tests of each combination, a generalized router is trained on the samples with all the tests, and a specialized router sends each combination to its own expert. We follow the released code. \textbf{REMIND} \citep{wu2026remind} adds to the soft routing a residual routing matrix for each group, which is used when the routing of the shared matrix is uncertain, and it trains the groups with a group DRO update. Its code is not released, so we reimplemented it from the description of the paper.

All the baselines use the same splits, seeds, features, class weights, batches, epochs and reported epoch as our framework, and they are run with the hyperparameters of their authors ($\gamma = 0.02$ for REMIND); Table~\ref{tab:main} gives their parameters. Flex-MoE uses the defaults of its released code ($16$ experts, $d = 128$). REMIND and Reweigh share the backbone of the REMIND paper: $128$ experts on EMBED, as in that paper, and $4$ experts with latent dimension $64$ on the tabular datasets, for which the paper gives no setting. Since they define a group by the combination of tests of its patients, they run unchanged when the groups have disjoint feature spaces. The excess loss of every method uses the references of our framework.

\section{Latent Space of the Three Datasets}
\label{app:latent_space}

Figure~\ref{fig:latent_appx} shows the latent space of the test patients of the three datasets, NHANES, Fed-Heart and EMBED, without and with the anchors. The $W_2$ distances and the nearest-anchor accuracies are in Table~\ref{tab:latent}. In NHANES and Fed-Heart, without the anchors each group lies in its own region of the latent space, whereas with the anchors the groups overlap and the two classes are organized around their anchors. In EMBED, the centroids of the groups coincide once the anchors are used, although the two largest groups, g4 and g6, still spread along different directions. The four density classes are ordered, and the four anchors lie close to each other at the scale of the latent space, so the figure shows them as nearly coincident.

\begin{figure*}[!htb]
\centering
\begin{subfigure}[t]{0.325\textwidth}
\centering
\includegraphics[width=\linewidth]{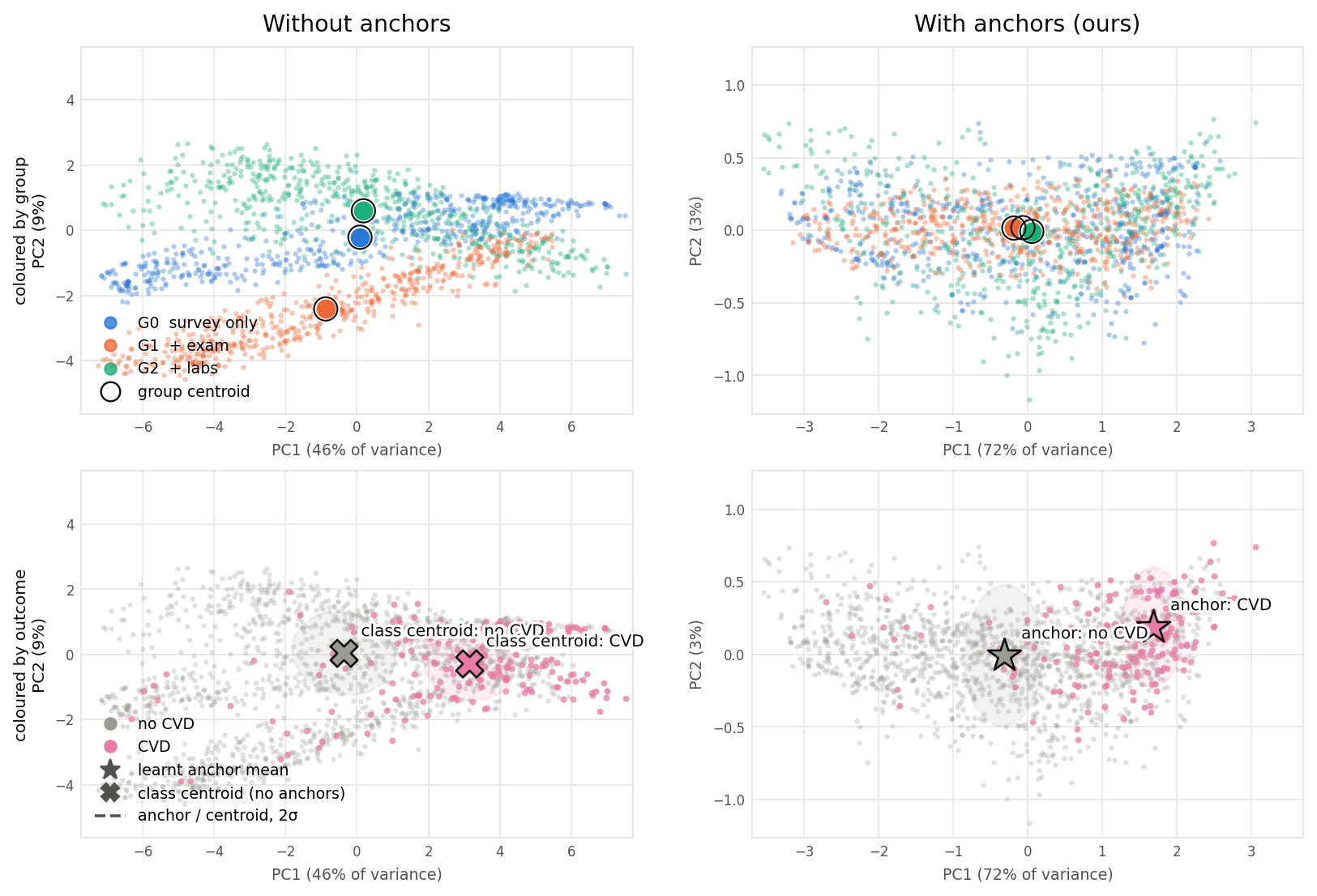}
\caption{NHANES, by group (top) and outcome (bottom).}
\label{fig:latent_nhanes}
\end{subfigure}
\hfill
\begin{subfigure}[t]{0.325\textwidth}
\centering
\includegraphics[width=\linewidth]{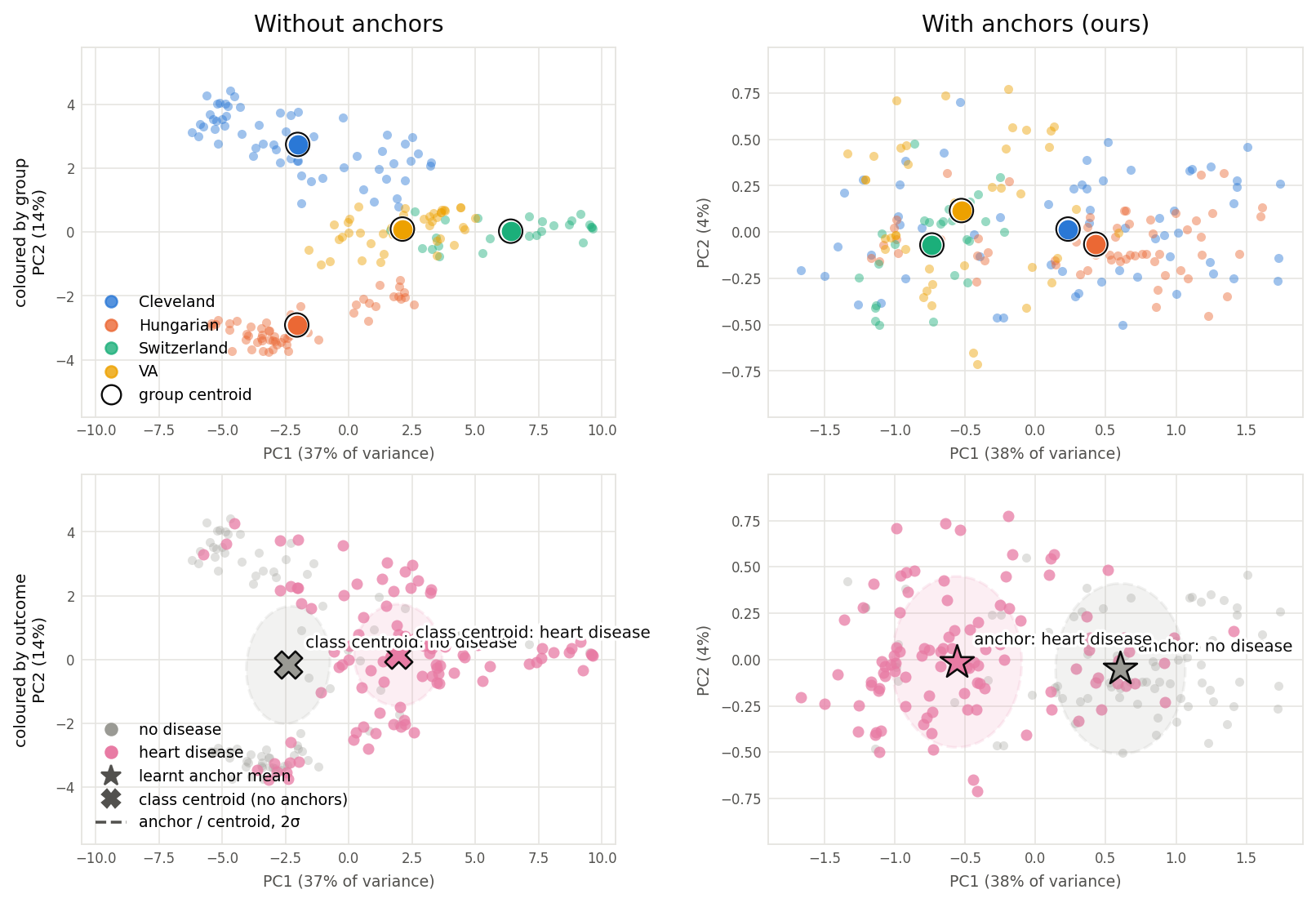}
\caption{Fed-Heart (fold 0), by hospital (top) and outcome (bottom).}
\label{fig:latent_fedheart}
\end{subfigure}
\hfill
\begin{subfigure}[t]{0.325\textwidth}
\centering
\includegraphics[width=\linewidth]{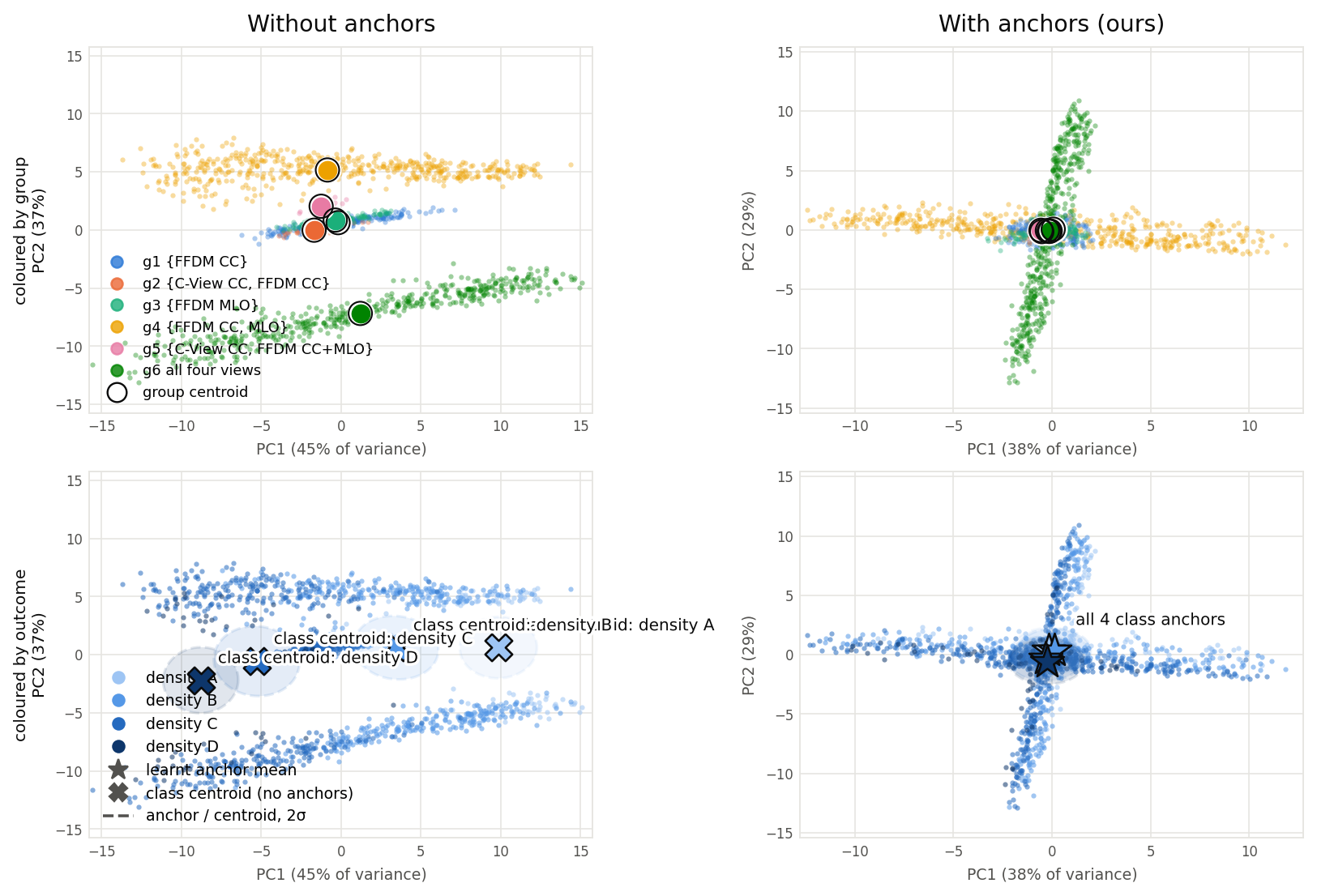}
\caption{EMBED, by group (top) and breast density (bottom).}
\label{fig:latent_embed}
\end{subfigure}
\caption{Latent space of the test patients of NHANES, Fed-Heart and EMBED without anchors (left panels) and with anchors (right panels). Each panel is projected on its first two principal components, and the axes are scaled per panel, since the anchors shrink the latent space. Rings are the centroids of the groups, stars the learnt anchors, and crosses the class centroids without anchors.}
\label{fig:latent_appx}
\end{figure*}

Beyond their effect on the losses (Appendix~\ref{app:ablations}), the anchors give the latent space two properties. First, they encourage groups to share the same space, as seen from the fact that the distance between groups drops from $0.53$, $1.08$ and $0.78$ to $0.03$, $0.33$ and $0.27$. This way a group will not end up being independent from the rest. Representations of this kind are used to transfer models between groups \citep{shen2018wasserstein}, and it is easier to reuse them for new tasks without revealing the group \citep{madras2018learning}. Second, they create a class organized latent space, which can be used to explain the final results. In this work, we do not evaluate these uses.

\section{Ablation Studies}
\label{app:ablations}

In the ablations, we examine what is the benefit of our framework's two main components: the alignment and separation losses, and the objective with respect to the groups. To evaluate every combination, we design the following ablation methods. 
\paragraph{Without the alignment and separation losses.} We set the weights $\alpha_{\mathrm{align}}$ and $\alpha_{\mathrm{sep}}$ of the final problem \eqref{eq:objective} to zero. Then the anchors are not used and nothing organizes the latent space, so the variants only differ in how they weight the groups.

\textbf{ERM} minimizes the average loss, where each group is weighted by its proportion $p_g$ in the training data and the weights are not updated:
\[
\min_{\phi,\,\psi}\ \sum_{g=1}^{G} p_g\, \Loss_g(\psi \circ \phi_g).
\]

\textbf{GroupDRO} maximizes over the weights of the groups, which optimizes the loss of the worst group. It is the final problem with $\alpha_{\mathrm{align}} = \alpha_{\mathrm{sep}} = 0$ and with the references set to zero:
\[
\min_{\phi,\,\psi}\ \max_{\lambda \in \Simplex}\ \sum_{g=1}^{G} \lambda_g\, \Loss_g(\psi \circ \phi_g).
\]

\textbf{Excess DRO} is the same, but each group is compared with its reference $\RefPt{g}$, i.e., it optimizes the excess loss of the worst group:
\[
\min_{\phi,\,\psi}\ \max_{\lambda \in \Simplex}\ \sum_{g=1}^{G} \lambda_g \big(\Loss_g(\psi \circ \phi_g) - \RefPt{g}\big).
\]

\paragraph{With the alignment and separation losses.} We use the weights $\alpha_{\mathrm{align}}, \alpha_{\mathrm{sep}} > 0$ of Appendix~\ref{app:training}, so the anchors are learned and the alignment loss pulls the groups towards them. The alignment loss of each group is weighted like its task loss.

\textbf{Anchors + ERM} adds the two losses to the ERM objective:
\[
\begin{aligned}
\min_{\phi,\,\psi,\,\mu}\ & \sum_{g=1}^{G} p_g \big(\Loss_g(\psi \circ \phi_g) + \alpha_{\mathrm{align}} \Loss^{\mathrm{align}}_g\big) \\
& + \alpha_{\mathrm{sep}} \Loss^{\mathrm{sep}}.
\end{aligned}
\]

\textbf{Anchors + GroupDRO} adds them to the group DRO objective, and the adversary now maximizes the task loss plus the alignment loss:
\[
\begin{aligned}
\min_{\phi,\,\psi,\,\mu}\ \Big\{ \max_{\lambda \in \Simplex}\ & \sum_{g=1}^{G} \lambda_g \big(\Loss_g(\psi \circ \phi_g) + \alpha_{\mathrm{align}} \Loss^{\mathrm{align}}_g\big) \\
& + \alpha_{\mathrm{sep}} \Loss^{\mathrm{sep}} \Big\}.
\end{aligned}
\]

\textbf{Anchors + Excess DRO} is our framework, i.e., the final problem \eqref{eq:objective} with the references $\RefPt{g}$ in place of the optimal losses:
\[
\begin{aligned}
\min_{\phi,\,\psi,\,\mu}\ \Big\{ \max_{\lambda \in \Simplex}\ & \sum_{g=1}^{G} \lambda_g \big(\Loss_g(\psi \circ \phi_g) - \RefPt{g} + \alpha_{\mathrm{align}} \Loss^{\mathrm{align}}_g\big) \\
& + \alpha_{\mathrm{sep}} \Loss^{\mathrm{sep}} \Big\}.
\end{aligned}
\]
All the variants have the same architecture as in the main paper, i.e., group-specific encoders and a shared head, and we test them in all the settings. The results are reported in Table~\ref{tab:ablation}.

\begin{center}
\begin{minipage}{\textwidth}
\centering
\captionof{table}{Results of the ablation studies in the original groups (left) and in the design with disjoint feature spaces (right). WG denotes the worst group, exc. the excess loss, i.e., the loss minus the reference, and $\Delta$ the difference between the largest and the smallest value of the metric across groups. Mean over $10$ seeds, best in bold. All the variants have the same number of parameters. No significance test is reported for these differences.}
\label{tab:ablation}
\footnotesize
\setlength{\tabcolsep}{2.2pt}
\resizebox{\textwidth}{!}{%
\begin{tabular}[t]{@{}c@{\hspace{2.5em}}c@{}}
\textbf{Original groups} & \textbf{Disjoint feature spaces}\\[2pt]
\begin{tabular}[t]{@{}lcccccc@{}}
\toprule
Variant & \shortstack{WG\\acc$\uparrow$} & \shortstack{$\Delta$\\acc$\downarrow$} & \shortstack{WG\\loss$\downarrow$} & \shortstack{$\Delta$\\loss$\downarrow$} & \shortstack{WG\\exc.$\downarrow$} & \shortstack{$\Delta$\\exc.$\downarrow$} \\
\midrule
\multicolumn{7}{@{}l}{\textit{NHANES}} \\
ERM & 60.2 & 12.2 & 0.607 & 0.155 & 0.328 & 0.118 \\
GroupDRO & 65.8 & 9.4 & 0.547 & 0.125 & 0.275 & 0.088 \\
Excess DRO & 64.3 & 10.1 & \textbf{0.541} & 0.114 & \textbf{0.264} & 0.078 \\
Anchors + ERM & 61.8 & 10.2 & 0.599 & 0.137 & 0.325 & 0.100 \\
Anchors + GroupDRO & \textbf{67.0} & \textbf{7.7} & 0.544 & 0.115 & 0.278 & 0.093 \\
Anchors + Excess DRO (ours) & 65.7 & 8.5 & 0.548 & \textbf{0.109} & 0.280 & \textbf{0.073} \\
\midrule
\multicolumn{7}{@{}l}{\textit{Fed-Heart}} \\
ERM & 74.6 & 18.9 & 0.579 & 0.274 & 0.230 & 0.129 \\
GroupDRO & \textbf{75.0} & \textbf{18.5} & 0.589 & 0.316 & \textbf{0.197} & \textbf{0.099} \\
Excess DRO & 74.5 & 19.0 & 0.558 & \textbf{0.207} & 0.276 & 0.187 \\
Anchors + ERM & 74.4 & 19.1 & \textbf{0.556} & 0.249 & 0.231 & 0.140 \\
Anchors + GroupDRO & 74.3 & 19.2 & 0.560 & 0.279 & 0.206 & 0.104 \\
Anchors + Excess DRO (ours) & 74.4 & 19.1 & 0.561 & 0.280 & 0.206 & 0.103 \\
\midrule
\multicolumn{7}{@{}l}{\textit{EMBED}} \\
ERM & 63.5 & 14.0 & 3.485 & 2.977 & 2.587 & 2.688 \\
GroupDRO & 63.8 & 14.1 & \textbf{1.049} & 0.538 & 0.177 & \textbf{0.252} \\
Excess DRO & \textbf{65.5} & \textbf{12.3} & 1.057 & 0.548 & \textbf{0.165} & 0.258 \\
Anchors + ERM & 56.9 & 20.3 & 1.456 & 0.941 & 0.565 & 0.650 \\
Anchors + GroupDRO & 62.6 & 13.4 & 1.077 & 0.557 & 0.247 & 0.328 \\
Anchors + Excess DRO (ours) & 62.2 & 14.0 & 1.052 & \textbf{0.534} & 0.240 & 0.318 \\
\bottomrule
\end{tabular}
&
\begin{tabular}[t]{@{}lcccccc@{}}
\toprule
Variant & \shortstack{WG\\acc$\uparrow$} & \shortstack{$\Delta$\\acc$\downarrow$} & \shortstack{WG\\loss$\downarrow$} & \shortstack{$\Delta$\\loss$\downarrow$} & \shortstack{WG\\exc.$\downarrow$} & \shortstack{$\Delta$\\exc.$\downarrow$} \\
\midrule
\multicolumn{7}{@{}l}{\textit{NHANES (designed)}} \\
ERM & 55.6 & 16.6 & 0.689 & 0.216 & 0.366 & 0.143 \\
GroupDRO & 62.8 & 10.0 & 0.585 & \textbf{0.098} & 0.287 & 0.054 \\
Excess DRO & 62.3 & 12.1 & 0.608 & 0.178 & 0.295 & 0.106 \\
Anchors + ERM & 60.8 & 9.9 & 0.628 & 0.126 & 0.317 & 0.054 \\
Anchors + GroupDRO & \textbf{65.6} & \textbf{5.6} & \textbf{0.581} & 0.111 & 0.284 & 0.067 \\
Anchors + Excess DRO (ours) & 64.5 & 7.4 & 0.590 & 0.123 & \textbf{0.278} & \textbf{0.050} \\
\midrule
\multicolumn{7}{@{}l}{\textit{Fed-Heart}} \\
ERM & 63.7 & 29.1 & \textbf{0.627} & 0.402 & 0.198 & 0.141 \\
GroupDRO & 63.8 & 29.7 & 0.628 & 0.412 & \textbf{0.189} & \textbf{0.132} \\
Excess DRO & \textbf{64.0} & \textbf{28.8} & \textbf{0.627} & 0.400 & 0.200 & 0.143 \\
Anchors + ERM & 62.9 & 30.0 & 0.632 & \textbf{0.339} & 0.266 & 0.205 \\
Anchors + GroupDRO & 63.5 & 29.3 & 0.635 & 0.385 & 0.223 & 0.159 \\
Anchors + Excess DRO (ours) & 63.4 & 29.4 & 0.635 & 0.389 & 0.219 & 0.154 \\
\midrule
\multicolumn{7}{@{}l}{\textit{EMBED}} \\
ERM & \textbf{63.3} & \textbf{10.2} & 3.217 & 2.623 & 2.385 & 2.556 \\
GroupDRO & 60.0 & 13.0 & \textbf{1.004} & \textbf{0.411} & \textbf{0.171} & \textbf{0.343} \\
Excess DRO & 59.5 & 13.5 & 1.021 & 0.428 & 0.189 & 0.361 \\
Anchors + ERM & 58.0 & 15.4 & 1.241 & 0.630 & 0.408 & 0.561 \\
Anchors + GroupDRO & 58.7 & 14.6 & 1.036 & 0.427 & 0.203 & 0.358 \\
Anchors + Excess DRO (ours) & 58.0 & 15.3 & 1.051 & 0.442 & 0.219 & 0.375 \\
\bottomrule
\end{tabular}
\end{tabular}}
\end{minipage}
\end{center}

\paragraph{Results.} Since all the variants share the architecture and differ only in the objective, the ablations test two predictions: the anchors should help where some groups cannot learn from their own data, and the excess loss should help where the references differ across groups and are reliable. The results support them only in part: our framework has the lowest worst-group excess loss only in NHANES (designed) ($0.278$ against $0.284$ for the next best variant); elsewhere it is within $0.009$ to $0.075$ of the best variant, and the comparison between the variants shows which condition is missing.
\newline
The anchors can help when there are groups that cannot train a good model on their own data. In the case of ERM, Anchors + ERM reduces the worst-group excess loss in cases where ERM is far worse than group DRO, as in EMBED ($2.587$ to $0.565$) and NHANES (designed) ($0.366$ to $0.317$). With group DRO, which already prioritizes the worst group, the anchors do not provide a clear benefit (the excess loss increases in five of the six settings). With excess DRO, they help in Fed-Heart ($0.276$ to $0.206$) and NHANES (designed) ($0.295$ to $0.278$). The excess loss moves the weight towards the groups that are furthest from their references, and this only helps if these groups can improve with the help of the others. In the other four settings, excess DRO without anchors is already the best variant or within $0.018$ of it, and the anchors increase its excess loss (by $0.016$ to $0.075$). This is not always the case: in the synthetic experiments with extreme imbalance, the rarest group cannot train a good model on its own, and the anchors do not change the worst-group regret in a significant way (Appendix~\ref{app:synthetic}). We do not evaluate the uses of the latent space mentioned in Appendix~\ref{app:latent_space}. However, we consider that an increase of at most $0.075$ in the excess loss does not justify giving up a latent space that is shared by the groups and organized by class.
\newline
The excess loss helps when it changes which group is the worst. This happens when the group with the highest loss is not the group that is furthest from its reference, which requires references that differ across groups and that are estimated reliably (Table~\ref{tab:refvalues}). In the synthetic experiments, where the optimal losses are known and spread, the excess loss clearly improves over group DRO (Appendix~\ref{app:synthetic}). In the real datasets, the references of NHANES are close ($0.242$ to $0.279$), and the two objectives are alike. In Fed-Heart they differ more, but those of the smallest groups are unreliable (the margin lowers the reference of Switzerland from $0.238$ to $0.076$), and excess DRO is clearly worse than group DRO ($0.276$ against $0.197$). NHANES (designed), where the references differ ($0.251$ to $0.323$) and the margins are small ($0.016$ to $0.034$), is where our framework is the best variant ($0.278$ against $0.284$).

\begin{center}
\begin{minipage}{\linewidth}
\centering
\captionof{table}{Estimated references per group. $\Est{g}$ is the raw estimate, $c_g$ the bootstrap margin and $\RefPt{g}$ the reference used in training. Fed-Heart: mean over the five outer folds. EMBED: no margin.}
\label{tab:refvalues}
\footnotesize
\setlength{\tabcolsep}{12pt}
\begin{tabular}{@{}llccc@{}}
\toprule
Setting & Group & $\Est{g}$ & $c_g$ & $\RefPt{g}$ \\
\midrule
\multirow{3}{*}{NHANES} & G0 & 0.310 & 0.031 & 0.279 \\
 & G1 & 0.272 & 0.029 & 0.242 \\
 & G2 & 0.266 & 0.014 & 0.252 \\
\midrule
\multirow{3}{*}{NHANES (designed)} & G0 & 0.357 & 0.034 & 0.323 \\
 & G1 & 0.281 & 0.030 & 0.251 \\
 & G2 & 0.301 & 0.016 & 0.285 \\
\midrule
\multirow{4}{*}{Fed-Heart} & Cleveland & 0.466 & 0.084 & 0.382 \\
 & Hungarian & 0.486 & 0.110 & 0.376 \\
 & Switzerland & 0.238 & 0.163 & 0.076 \\
 & VA & 0.567 & 0.146 & 0.421 \\
\midrule
\multirow{4}{*}{Fed-Heart (disjoint)} & Cleveland & 0.624 & 0.054 & 0.570 \\
 & Hungarian & 0.457 & 0.107 & 0.349 \\
 & Switzerland & 0.237 & 0.225 & 0.027 \\
 & VA & 0.567 & 0.098 & 0.468 \\
\midrule
\multirow{6}{*}{EMBED} & g1 & 0.685 & --- & 0.685 \\
 & g2 & 0.702 & --- & 0.702 \\
 & g3 & 0.767 & --- & 0.767 \\
 & g4 & 0.624 & --- & 0.624 \\
 & g5 & 0.897 & --- & 0.897 \\
 & g6 & 0.587 & --- & 0.587 \\
\midrule
\multirow{4}{*}{EMBED (disjoint)} & g1 & 0.694 & --- & 0.694 \\
 & g2 & 0.832 & --- & 0.832 \\
 & g3 & 0.756 & --- & 0.756 \\
 & g6 & 0.765 & --- & 0.765 \\
\bottomrule
\end{tabular}
\end{minipage}
\end{center}

\section{Synthetic Experiments}
\label{app:synthetic}

We use synthetic data to test the objective in a setting where the optimal loss of every group is known up to a small Monte Carlo error. 

\subsection{Data Description}
\label{sec:synthetic_data}
We create the data so that all the patients are drawn from the same distribution, and the groups only differ in which tests they record. There are four tests in the shared tests structure and six in the disjoint tests structure (described below), and each test consists of three features. A patient is generated in three steps. First, the class is drawn uniformly, $y\sim\mathrm{Unif}\{1,2,3\}$. Second, a latent vector $z\in\mathbb{R}^4$, which is never observed and whose dimension does not depend on the number of tests, is drawn around the mean of the class,
\[
z\mid y\sim\mathcal{N}(m_y, I_4).
\]
The class means $m_1,m_2,m_3$ are drawn once from $\mathcal{N}(0, 3^2 I_4)$. Third, the value of every test is computed from the same latent vector. Test $v$ is a noisy linear view of $z$,
\[
u_v = A_v z + \sigma_v \varepsilon_v \in\mathbb{R}^3, \qquad \varepsilon_v\sim\mathcal{N}(0, I_3),
\]
followed by the element-wise map
\[
x_v = u_v + \tfrac{1}{2}\sin u_v,
\]
where $A_v\in\mathbb{R}^{3\times 4}$ is a fixed matrix with independent $\mathcal{N}(0,1/4)$ entries, $\sigma_v$ is the noise level of the test, and the noises $\varepsilon_v$ are independent across tests and patients.
\newline
A group $g$ is defined by a subset $S_g$ of the tests. Its patients are independent draws from the process above, and only the tests in $S_g$ are recorded, so the features of a patient of group $g$ are $x_{S_g}=(x_v)_{v\in S_g}$. All the groups sample from the same joint distribution of the class and the tests, and they differ only in which tests they record. The class means and the matrices $A_v$ are drawn once, and they are the same in all the seeds; the seeds change the samples and the initialization of the models.

\textbf{Optimal loss.} Since the process is known, the optimal loss of a group can be computed. For the log loss, no predictor has an expected loss below the conditional entropy of the class given the tests $S_g$, which the true posterior attains \citep{cover2006elements}, so
\[
R^\star_g = \mathbb{E}\big[-\log \Prob(y\mid x_{S_g})\big].
\]
This is the optimal loss over all predictors, so it lower-bounds the optimal loss of the model class, and the regret that we report includes the approximation error of the model. The map from $u_v$ to $x_v$ is invertible, so we can condition on $u_{S_g}$. Let $A_{S_g}$ be the matrix that stacks the $A_v$ with $v\in S_g$, and $D_{S_g}$ the diagonal matrix with the entries $\sigma_v^2$, repeated three times for each test. Then $u_{S_g}\mid y\sim\mathcal{N}(A_{S_g}m_y,\Sigma_g)$ with $\Sigma_g=A_{S_g}A_{S_g}^\top+D_{S_g}$, the same for every class, so the posterior is a softmax,
\[
\Prob(y=c\mid u_{S_g}) = \operatorname{softmax}_c\Big(-\tfrac12\,(u_{S_g}-\mu_{g,c})^\top\Sigma_g^{-1}(u_{S_g}-\mu_{g,c})\Big),\qquad \mu_{g,c}=A_{S_g}m_c.
\]
We estimate $R^\star_g$ by averaging $-\log\Prob(y\mid u_{S_g})$ over $200{,}000$ samples of the process, with a standard error below $0.002$, and the Bayes accuracy in the same way. A test makes the class easier to predict when its noise is small, but also depends on its matrix $A_v$, so the optimal loss does not follow the noise levels alone.

We simulate two structures in the experiments.  1) The \emph{shared tests}, in which we have four tests and the groups record $\{0,1\}$, $\{0,2\}$, $\{0,3\}$ and $\{0,1,2,3\}$. Test $0$ is recorded by every group. The noise is $\sigma=(8, 8, 1.6, 1.0)$. 2) The \emph{disjoint tests}, in which we have six tests and the groups record $\{0,1\}$, $\{2,3\}$ and $\{4,5\}$, so no test is shared. The noise is $\sigma=(8, 8, 2, 2, 0.9, 0.9)$.

\textbf{Widely spread optimal losses.} The optimal losses are $(1.01, 0.33, 0.40, 0.15)$ in the shared tests structure and $(1.01, 0.27, 0.12)$ in the disjoint one, while a constant predictor has loss $\log 3 = 1.10$ (Figure~\ref{fig:synthetic_groups}). The largest group only records very noisy tests, so its optimal loss is close to chance, and the rarest group records the clean tests. We chose the noise values from the optimal losses alone, before training any model, to make their spread large ($0.86$ and $0.89$). 

\textbf{Group sizes.} In the shared tests structure we use two sets of sizes, $(4000, 1000, 250, 100)$ and an extreme imbalance $(20000, 2000, 200, 30)$. In the disjoint tests structure the sizes are $(3000, 3000, 100)$ and $(3000, 3000, 30)$. In all of them the largest group is the least informative and the rarest group is the most informative. The validation set of a group has $25\%$ of its training size (at least $20$ samples), and the test set has $2000$ samples.

\textbf{Methods and metric.} All the methods use the architecture of the main paper: one encoder per group (an MLP with two hidden layers of $32$ units and latent dimension $8$), a shared linear head, and, when anchors are used, one diagonal Gaussian anchor per class with $\alpha_{\mathrm{align}}=\alpha_{\mathrm{sep}}=0.1$. The methods are the six variants of the ablation studies (Appendix~\ref{app:ablations}): ERM, GroupDRO, Excess DRO, each with and without anchors, where Anchors + Excess DRO is our framework. We also train a dedicated model per group, with its own encoder and head. We train with Adam for $1000$ steps on minibatches with the proportions of the groups, we update the group weights every $10$ steps with step size $0.1$, and we estimate the references with $3$-fold cross-fitting and the correction in Eq.~\eqref{eq:refpt}. We use the signed excess loss and detach the head in the separation loss, as in the main experiments, and we keep the checkpoint with the lowest worst-group validation loss. The metric is the \emph{worst-group regret}, $\max_g(\text{test loss}_g - R^\star_g)$, where $R^\star_g$ is the optimal loss of the data-generating process (Section~\ref{sec:synthetic_data}) and not the estimated reference. We report the mean and the standard deviation over $6$ seeds.

\begin{figure*}[t]
\centering
\includegraphics[width=0.85\textwidth]{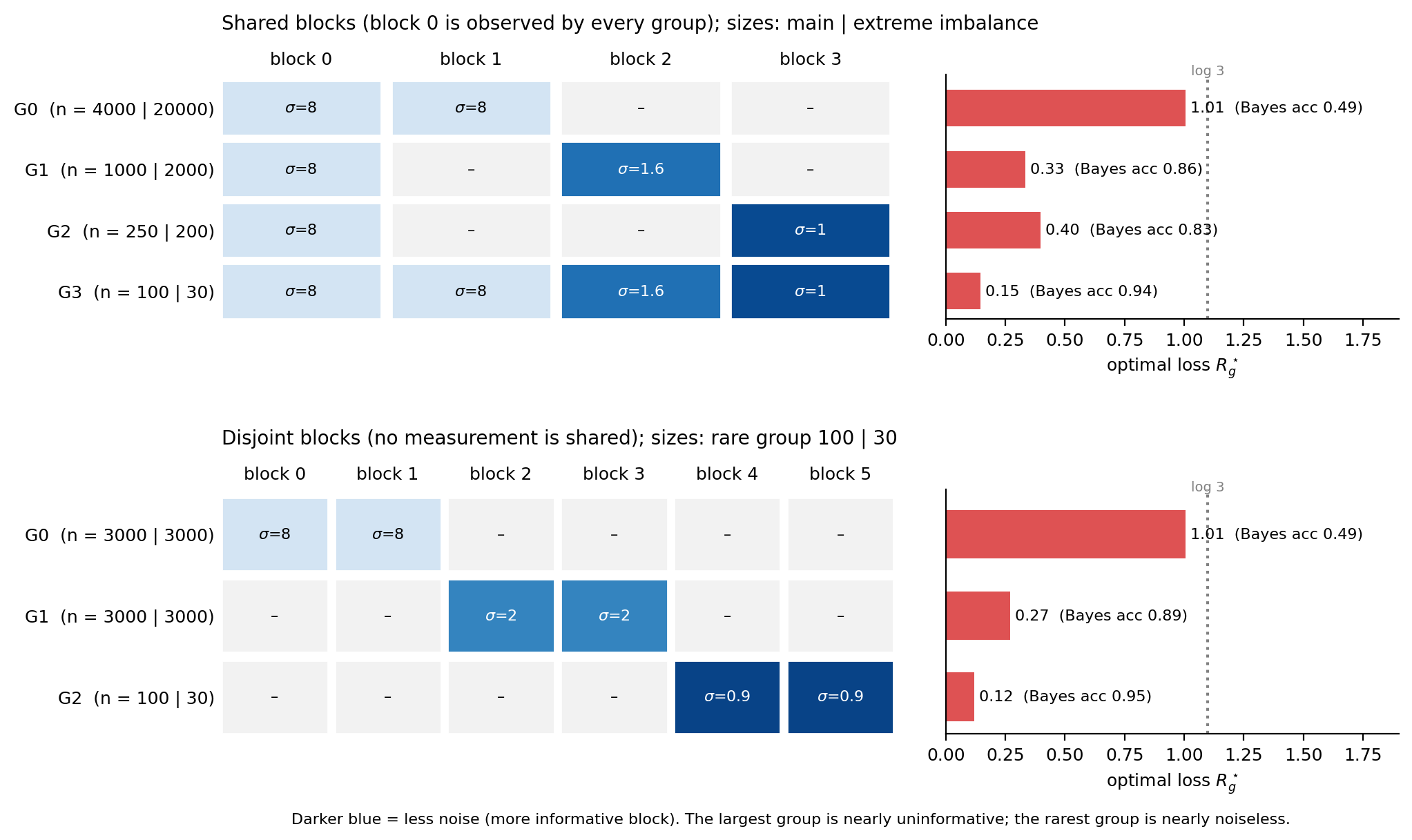}
\caption{The groups of the synthetic experiments. Left: the tests that each group records, with the noise $\sigma$ of each test (darker is less noisy) and the training size of each group for the two sets of sizes. Right: the optimal loss $R^\star_g$ of each group and its Bayes accuracy. The dotted line is the loss of a constant predictor.}
\label{fig:synthetic_groups}
\end{figure*}

\subsection{Results and Discussion}
\label{sec:synthetic_results}

\begin{table}[t]
\caption{Worst-group regret on the synthetic data (lower is better): mean $\pm$ standard deviation over $6$ seeds, best in bold. Shared tests: main sizes and extreme imbalance; disjoint tests: rare group of $100$ and of $30$ samples.}
\label{tab:synthetic}
\centering
\footnotesize
\setlength{\tabcolsep}{3pt}
\resizebox{\columnwidth}{!}{%
\begin{tabular}{@{}lcccc@{}}
\toprule
 & \multicolumn{2}{c}{Shared tests} & \multicolumn{2}{c}{Disjoint tests} \\
\cmidrule(lr){2-3}\cmidrule(lr){4-5}
Method & main & imbalance & rare $100$ & rare $30$ \\
\midrule
Dedicated & 0.435 $\pm$ 0.085 & 0.831 $\pm$ 0.069 & 0.124 $\pm$ 0.034 & 0.196 $\pm$ 0.038 \\
ERM & 0.531 $\pm$ 0.166 & 0.929 $\pm$ 0.040 & 0.289 $\pm$ 0.123 & 0.620 $\pm$ 0.197 \\
GroupDRO & 0.508 $\pm$ 0.161 & 0.672 $\pm$ 0.115 & 0.270 $\pm$ 0.096 & 0.460 $\pm$ 0.129 \\
Excess DRO & 0.387 $\pm$ 0.088 & 0.653 $\pm$ 0.162 & 0.124 $\pm$ 0.049 & \textbf{0.195} $\pm$ 0.062 \\
\midrule
Anchors + ERM & 0.639 $\pm$ 0.066 & 0.924 $\pm$ 0.053 & 0.434 $\pm$ 0.164 & 0.677 $\pm$ 0.186 \\
Anchors + GroupDRO & 0.616 $\pm$ 0.114 & \textbf{0.628} $\pm$ 0.048 & 0.291 $\pm$ 0.142 & 0.380 $\pm$ 0.101 \\
\textbf{Anchors + Excess DRO (ours)} & \textbf{0.355} $\pm$ 0.055 & 0.637 $\pm$ 0.099 & \textbf{0.116} $\pm$ 0.023 & 0.210 $\pm$ 0.053 \\
\bottomrule
\end{tabular}}
\end{table}

\textbf{Effect of the objective.} ERM follows the proportions of the groups, so it is dominated by the largest group, which is almost at its optimal loss, and it sacrifices performance for the rare groups: it is worse than training each group separately in all the settings. Group DRO on the raw loss helps little when the imbalance is moderate, since the group with the highest loss is the largest one and the one whose loss cannot be decreased. In the main sizes, it leaves the rarest group with a regret of $0.51$ while the largest group is at $0.02$. The excess loss selects instead the group that is furthest from its optimum, the rarest group, and it lowers its regret (to $0.39$ in the main sizes). In the extreme imbalance setting, the two objectives are similar, since the underrepresented group ends up being the one with the highest loss. In that case both excess DRO and Group DRO select it. In the setting with imbalance across groups is what makes ERM fail, and any objective that reweights the groups corrects for this.

\textbf{Effect of the anchors.} With Excess DRO, adding the anchors does not change the worst-group regret in a significant way in any setting, so, in this world, the shared latent space that they give costs nothing in prediction. This is not the case with the other objectives: with ERM the anchors tend to increase the regret, and with group DRO on the raw loss the effect has no consistent sign. The anchors are therefore safe to use together with the excess loss, and the worst-group regret alone does not separate the variants with and without them. In this setting, every group has its own encoder and all the groups share the same relation between the latent vector and the class, so aligning the groups is not necessary to predict well, and the results do not show a benefit of the anchors, including for the rarest group under the extreme imbalance. The properties of the latent space that the anchors give are analyzed on the real datasets in Appendix~\ref{app:latent_space}.

\textbf{Effect of sharing.} A single model with a shared head can serve groups that share no test without losing performance, provided that it is trained with the excess loss. In the disjoint tests structure, where the groups can only share information through the head and the latent space, our framework is comparable to the dedicated models, whereas ERM and group DRO on the raw loss, with and without anchors, are worse than them. In the shared tests structure, our framework is better than the dedicated models, with the main sizes ($0.36$ against $0.44$) and, more clearly, with the extreme imbalance ($0.64$ against $0.83$), probably because the rarest group, with only $30$ samples, benefits from the shared head.
\section{Proof of Proposition~\ref{prop:order}}
\label{app:regret}
Let $\Xcal_g$ be the feature space of group $g$, and let $\Xcal_h=\Xcal_g\times\Xcal'$ be the feature space of group $h$, so that a patient of group $h$ has covariates $x_h=(x_g,x')$, where $x'$ are the extra tests. For predictors $f_g\in\Fcal_g$ and $f_h\in\Fcal_h$, the losses of the two groups are
\begin{align*}
\Loss_g(f_g)&=\E_{(x,y)\sim\Prob^g_{X,Y}}\big[\Loss(f_g;x,y)\big],\\
\Loss_h(f_h)&=\E_{(x,y)\sim\Prob^h_{X,Y}}\big[\Loss(f_h;x,y)\big],
\end{align*}
and their optimal losses are $\Ach{g}=\inf_{f_g\in\Fcal_g}\Loss_g(f_g)$ and $\Ach{h}=\inf_{f_h\in\Fcal_h}\Loss_h(f_h)$.
\newline
The model class of $h$ contains the predictors that ignore the extra tests. That is, for every $f_g\in\Fcal_g$, there is a predictor $f_{h\to g}(x_g,x'):=f_g(x_g)$ that belongs to $\Fcal_h$. For such an $f_h$, we show that $\Loss_h(f_h)=\Loss_g(f_g)$. \newline
The only assumption is that groups $g$ and $h$ describe the same population: if $(x,y)\sim\Prob^h_{X,Y}$ with $x=(x_g,x')$, then $(x_g,y)\sim\Prob^g_{X,Y}$. In other words, the distribution of group $g$ is the distribution of group $h$ after removing the extra tests.
\newline
We start from the loss of $h$ and write it as an expectation over the distribution of $h$:
\begin{align*}
\Loss_h(f_{h\to g})
&=\E_{(x,y)\sim\Prob^h_{X,Y}}\big[\Loss(f_{h\to g};x,y)\big]\\
&=\E_{(x,y)\sim\Prob^h_{X,Y}}\big[\Loss(f_g;x_g,y)\big]\\
&=\E_{(x_g,y)\sim\Prob^g_{X,Y}}\big[\Loss(f_g;x_g,y)\big]\\
&=\Loss_g(f_g).
\end{align*}
The first equality is the definition of $\Loss_h$. In the second, we replace $\Loss(f_{h\to g};x,y)$ by $\Loss(f_g;x_g,y)$, since both predictors give the same output, $f_{h\to g}(x_g,x')=f_g(x_g)$, and the loss of a predictor at $(x,y)$ only depends on its output and on $y$ (as for the cross-entropy). The expectation is still over the distribution of $h$, but the quantity inside it is now a function of $(x_g,y)$ only. In the third equality, we use the assumption: when $(x,y)\sim\Prob^h_{X,Y}$, the pair $(x_g,y)$ has distribution $\Prob^g_{X,Y}$, so the expectation of a function of $(x_g,y)$ under $\Prob^h_{X,Y}$ is its expectation under $\Prob^g_{X,Y}$. The last equality is the definition of $\Loss_g$.
 \newline
Finally, $\Ach{h}=\inf_{f_h\in\Fcal_h}\Loss_h(f_h)$ and $f_{h\to g}\in\Fcal_h$, so
\[
\Ach{h}\;\le\;\Loss_h(f_{h\to g})\;=\;\Loss_g(f_g)\qquad\text{for every }f_g\in\Fcal_g.
\]
Taking the infimum over $f_g\in\Fcal_g$ on the right-hand side gives
\[
\Ach{h}\;\le\;\inf_{f_g\in\Fcal_g}\Loss_g(f_g)\;=\;\Ach{g}.\hfill\square
\]

\section{Proof of Proposition~\ref{prop:separate}}
\label{app:separate}

Let $\phi_g\in\Phi_g$ and $\psi_g\in\Psi$ be the encoder and the head of group $g$, for $g=1,\dots,G$, so that group $g$ predicts with $\psi_g\circ\phi_g$ when it has its own head. We want to find encoders $\tilde\phi_g$ and a single head $\tilde\psi$ that give the same prediction as $\psi_g\circ\phi_g$ for every patient of every group.
\newline
The idea is to move the representations of the groups to different regions of $\Zcal$. Then the head can recognize the group of a patient from the position of its representation, and it applies the head of that group.
\newline
We formalize the assumption that the encoder of each group can place its representation anywhere in $\Zcal$ with two properties: the outputs of the encoders are bounded, i.e., $\sup_{x\in\Xcal_g}\|\phi_g(x)\|<\infty$, and adding a constant vector to an encoder of $\Phi_g$ gives an encoder of $\Phi_g$ (for example, through the bias of its last layer). We also assume that $\Psi$ is the set of all the functions from $\Zcal$ to $\R^C$.
\newline
\textbf{Step 1: bounded representations.} Since there are finitely many groups and the encoders are bounded, there is $R>0$ such that
\[
\|\phi_g(x)\|\le R\qquad\text{for every group }g\text{ and every }x\in\Xcal_g.
\]
That is, all the representations of all the groups lie in the ball of radius $R$ around the origin, so the groups can overlap there.
\newline
\textbf{Step 2: moving the groups apart.} Let $e_1$ be the first vector of the canonical basis of $\Zcal=\R^k$, and let $a_g:=3gR\,e_1$. For two different groups,
\[
\|a_g-a_h\|=3R\,|g-h|\ge3R>2R,
\]
so the centers $a_g$ are more than $2R$ apart. We shift the encoder of each group,
\[
\tilde\phi_g:=\phi_g+a_g,
\]
which is an encoder of $\Phi_g$ by the assumption on the shifts. Let $B_g:=\{z\in\Zcal:\|z-a_g\|\le R\}$ be the ball of radius $R$ around $a_g$. For every $x\in\Xcal_g$,
\[
\|\tilde\phi_g(x)-a_g\|=\|\phi_g(x)\|\le R,
\]
so the new representations of group $g$ lie in $B_g$. They are the old ones moved by $a_g$.
\newline
\textbf{Step 3: the regions do not overlap.} Suppose that $z\in B_g\cap B_h$ for two different groups. By the triangle inequality,
\[
\|a_g-a_h\|\le\|a_g-z\|+\|z-a_h\|\le2R,
\]
which contradicts $\|a_g-a_h\|>2R$. Therefore every point of $\Zcal$ belongs to at most one of the balls $B_1,\dots,B_G$.
\newline
\textbf{Step 4: the shared head.} We define the head by what each group needs in its own ball,
\[
\tilde\psi(z)=\begin{cases}\psi_g(z-a_g) & \text{if } z\in B_g,\\ 0 & \text{if } z\notin\bigcup_{g}B_g.\end{cases}
\]
For $z\in B_g$, the point $z-a_g$ is in the ball of radius $R$ around the origin, where the representations of group $g$ were before the shift, so $\psi_g$ is applied to a point of its own group. The definition is unambiguous because, by Step 3, $z$ cannot be in two balls. It is a function from $\Zcal$ to $\R^C$, so it is a head of $\Psi$. The value $0$ outside the balls is arbitrary, since no patient is represented there.
\newline
\textbf{Step 5: same predictions.} Let $x\in\Xcal_g$. By Step 2, $\tilde\phi_g(x)\in B_g$, so
\begin{align*}
\tilde\psi\big(\tilde\phi_g(x)\big)&=\psi_g\big(\tilde\phi_g(x)-a_g\big)\\
&=\psi_g\big(\phi_g(x)\big).
\end{align*}
The first equality is the definition of $\tilde\psi$ on $B_g$, and the second is the definition of $\tilde\phi_g$. Since $x$ and $g$ are arbitrary, $\tilde\psi\circ\tilde\phi_g=\psi_g\circ\phi_g$ for every group $g$.\hfill$\square$

% ======================================================================================
% SECTION 1: theory of the anchors and of the latent space losses
% ======================================================================================
\section{Class Anchors and Latent Space Losses}
\label{app:losses}
Each class $c$ has a Gaussian anchor $\mu_c=\mathcal{N}(m_c,\Sigma_c)$ on the latent space $\Zcal=\R^k$. We restrict the covariance to be diagonal, $\Sigma_c=\operatorname{diag}(v_c)$ with variances $v_c\in\R^k_{>0}$. The means $m_c$ and the variances $v_c$ are learned together with the encoders and the head. 
\subsection{The 2-Wasserstein Distance}
Let $\nu$ and $\mu$ be two distributions on $\R^k$ with finite second moments, i.e., $\E_{a\sim\nu}\|a\|_2^2<\infty$ and $\E_{b\sim\mu}\|b\|_2^2<\infty$, so that their means and covariances exist. The 2-Wasserstein distance between them is
\[
\Wtwo^2(\nu,\mu)=\inf_{\pi\in\Pi(\nu,\mu)}\E_{(a,b)\sim\pi}\|a-b\|_2^2,
\]
where $\Pi(\nu,\mu)$ is the joint distributions of a pair $(a,b)$ whose marginals are $\nu$ and $\mu$.

For two Gaussians it has a closed form \citep{dowson1982frechet,gelbrich1990formula,bhatia2019bures},
\[
\begin{aligned}
&\Wtwo^2\big(\mathcal{N}(m_1,\Sigma_1),\mathcal{N}(m_2,\Sigma_2)\big)\\
&\qquad=\|m_1-m_2\|_2^2+\operatorname{tr}(\Sigma_1+\Sigma_2)\\
&\qquad\quad-2\operatorname{tr}\Big(\big(\Sigma_2^{1/2}\Sigma_1\Sigma_2^{1/2}\big)^{1/2}\Big),
\end{aligned}
\]
where $\operatorname{tr}(A)$ is the trace of a matrix $A$, defined as the sum of its diagonal entries, and $\Sigma^{1/2}$ is the square root of the covariance matrix $\Sigma$.

In this case both covariances are diagonal: \newline $\Sigma_i=\operatorname{diag}(v_{i,1},\dots,v_{i,k})$ with $v_{i,j}>0$. \newline Then $\Sigma_i^{1/2}=\operatorname{diag}(\sqrt{v_{i,1}},\dots,\sqrt{v_{i,k}})$ and for the components of the closed form distance we have \newline 
\[
\big(\Sigma_2^{1/2}\Sigma_1\Sigma_2^{1/2}\big)^{1/2}=\operatorname{diag}\big(\sqrt{v_{1,1}v_{2,1}},\dots,\sqrt{v_{1,k}v_{2,k}}\big).
\]
Given the trace definition, we then have \newline $\operatorname{tr}(\Sigma_1+\Sigma_2)=\sum_{j=1}^k(v_{1,j}+v_{2,j})$ and $\operatorname{tr}\big((\Sigma_2^{1/2}\Sigma_1\Sigma_2^{1/2})^{1/2}\big)=\sum_{j=1}^k\sqrt{v_{1,j}v_{2,j}}$. \newline Substituting them in the closed form gives
\[
\begin{aligned}
&\Wtwo^2\big(\mathcal{N}(m_1,\operatorname{diag}(v_1)),\mathcal{N}(m_2,\operatorname{diag}(v_2))\big)\\
&\qquad=\|m_1-m_2\|_2^2+\sum_{j=1}^k\Big(v_{1,j}+v_{2,j}-2\sqrt{v_{1,j}v_{2,j}}\Big).
\end{aligned}
\]
Since $\|m_1-m_2\|_2^2=\sum_{j=1}^k(m_{1,j}-m_{2,j})^2$ and \newline $v_{1,j}+v_{2,j}-2\sqrt{v_{1,j}v_{2,j}}=\big(\sqrt{v_{1,j}}-\sqrt{v_{2,j}}\big)^2$, we obtain
\begin{equation}
\begin{aligned}
&\Wtwo^2\big(\mathcal{N}(m_1,\operatorname{diag}(v_1)),\mathcal{N}(m_2,\operatorname{diag}(v_2))\big)\\
&\qquad=\sum_{j=1}^k\Big[(m_{1,j}-m_{2,j})^2+\big(\sqrt{v_{1,j}}-\sqrt{v_{2,j}}\big)^2\Big].
\end{aligned}
\label{eq:w2diag}
\end{equation}
\subsection{The Alignment Loss}

The class-conditional distribution $\nu_g^c$ of the embeddings is not Gaussian in general, and its distance to an anchor has no closed form. We therefore compare the anchor with the diagonal Gaussian that has the same mean and the same coordinate-wise variances as $\nu_g^c$,
\[
\begin{aligned}
\tilde\nu_g^c&=\mathcal{N}\big(m_g^c,\operatorname{diag}(v_g^c)\big),\\
m_g^c&=\E[\phi_g(X)\mid Y=c],\quad v_g^c=\operatorname{Var}[\phi_g(X)\mid Y=c],
\end{aligned}
\]
where the moments are taken under $\Prob^g_{X,Y}$. By \eqref{eq:w2diag}, $\Wtwo^2(\tilde\nu_g^c,\mu_c)$ is available in closed form, and it never exceeds the distance that it replaces, $\Wtwo^2(\nu_g^c,\mu_c)$, as shown in Lemma~\ref{lem:lower}.
\begin{lemma}\label{lem:lower}
Let $\nu$ be a distribution on $\R^k$ with finite second moments, mean $m$ and coordinate-wise variances $v\in\R^k_{>0}$, and let $\mu=\mathcal{N}(m',\operatorname{diag}(v'))$ with $v'\in\R^k_{>0}$. Let $\tilde\nu=\mathcal{N}(m,\operatorname{diag}(v))$ be the diagonal Gaussian with the same mean and coordinate-wise variances as $\nu$. Then
\[
\Wtwo^2(\tilde\nu,\mu)\;\le\;\Wtwo^2(\nu,\mu).
\]
\end{lemma}
\begin{proof}
Let $\pi\in\Pi(\nu,\mu)$ be any coupling of $\nu$ and $\mu$, and let $(a,b)\sim\pi$, so that $a\sim\nu$ and $b\sim\mu$. For each coordinate $j$,
\begin{align*}
\E[a_j]&=m_j, & \operatorname{Var}(a_j)&=v_j,\\
\E[b_j]&=m'_j, & \operatorname{Var}(b_j)&=v'_j,
\end{align*}
and all these quantities are finite since $\nu$ and $\mu$ have finite second moments.
\newline
We want to show that the cost of $\pi$ is at least $\Wtwo^2(\tilde\nu,\mu)$. The cost splits over the coordinates,
\[
\E_{(a,b)\sim\pi}\big[\|a-b\|_2^2\big]=\sum_{j=1}^k\E\big[(a_j-b_j)^2\big].
\]
For a coordinate $j$, we have,
\begin{align*}
\E\big[(a_j-b_j)^2\big]&=\big(\E[a_j-b_j]\big)^2+\operatorname{Var}(a_j-b_j)\\
&=(m_j-m'_j)^2+v_j+v'_j-2\operatorname{Cov}(a_j,b_j).
\end{align*}
By the Cauchy--Schwarz inequality,
\begin{align*}
\operatorname{Cov}(a_j,b_j)&=\E\big[(a_j-m_j)(b_j-m'_j)\big]\\
&\le\sqrt{\E\big[(a_j-m_j)^2\big]\;\E\big[(b_j-m'_j)^2\big]}\\
&=\sqrt{v_jv'_j},
\end{align*}
and therefore
\begin{align*}
\E\big[(a_j-b_j)^2\big]&\ge(m_j-m'_j)^2+v_j+v'_j-2\sqrt{v_jv'_j}\\
&=(m_j-m'_j)^2+\big(\sqrt{v_j}-\sqrt{v'_j}\big)^2.
\end{align*}
Going back to the cost of $\pi$, we replace each term of the sum by this lower bound:
\begin{align*}
\E_{(a,b)\sim\pi}\big[\|a-b\|_2^2\big]
&=\sum_{j=1}^k\E\big[(a_j-b_j)^2\big]\\
&\ge\sum_{j=1}^k\Big[(m_j-m'_j)^2\\
&\qquad\qquad+\big(\sqrt{v_j}-\sqrt{v'_j}\big)^2\Big]\\
&=\Wtwo^2(\tilde\nu,\mu),
\end{align*}
where the last equality is \eqref{eq:w2diag}. Since this holds for every coupling $\pi\in\Pi(\nu,\mu)$, taking the infimum over $\pi$ gives $\Wtwo^2(\nu,\mu)\ge\Wtwo^2(\tilde\nu,\mu)$.
\end{proof}
Lemma~\ref{lem:lower} shows that the moment loss in \eqref{eq:align} lower-bounds the corresponding full-distribution Wasserstein loss:
\begin{align*}
\Loss^{\mathrm{align}}_g&=\frac{1}{C}\sum_{c=1}^{C}\Wtwo^2\big(\tilde\nu_g^{c},\mu_c\big)\\
&\le\frac{1}{C}\sum_{c=1}^{C}\Wtwo^2\big(\nu_g^{c},\mu_c\big).
\end{align*}

\subsection{The Separation Loss}
The separation loss \eqref{eq:sep} is the average over the classes of the expected loss of the head on samples from the anchors. We use the cross-entropy loss \newline $\Loss(\psi;z,c)=-\log\operatorname{softmax}(\psi(z))_c$, and the separation loss then becomes
\begin{align*}
\Loss^{\mathrm{sep}}=\frac{1}{C}\sum_{c=1}^{C}\E_{z\sim\mu_c}\big[-\log\operatorname{softmax}(\psi(z))_c\big].
\end{align*}
The separation loss penalizes anchors that the head cannot classify, so it keeps the anchors distinct enough for the shared head to separate the classes. It does not impose a fixed distance between their means.
\section{Anchor Algorithm Implementation}
\label{app:anchor-impl}
This appendix describes how the anchors and the losses of Appendix~\ref{app:losses} are implemented in Algorithm~\ref{alg:train} in lines~\ref{alg:nu}--\ref{alg:align}.
\subsection{Parameterization and Initialization of the Anchors}
The means $m_c$ are trained directly. The variances must be positive, so we train an unconstrained vector $s_c\in\R^k$ and define
\[
v_c=\operatorname{softplus}(s_c),
\]
where $\operatorname{softplus}(s)=\log(1+e^{s})>0$ is applied to each entry. At initialization, on NHANES and Fed-Heart, $m_c=0$ and every entry of $s_c$ is $-3$, which gives a variance of $\operatorname{softplus}(-3)\approx0.05$ in every coordinate, for all the classes; on EMBED, the entries of $m_c$ are drawn from $\mathcal{N}(0,0.1^2)$ and $s_c$ is set so that the initial variance is $1$ in every coordinate. For numerical stability, the implementation adds a small constant to every variance, $10^{-4}$ in all the experiments, which we omit from the notation.
\newline
Samples from an anchor are drawn with the reparameterization trick,
\[
z_{c,j}=m_{c,j}+\sqrt{v_{c,j}}\,\xi_j,\quad j=1,\dots,k,\qquad \xi\sim\mathcal{N}(0,I_k).
\]
The sample is then a differentiable function of $m_c$ and $v_c$, so the losses can be differentiated with respect to the anchors and the anchors are updated by gradient steps.

\subsection{Minibatch Estimator of the Alignment Loss}
In a minibatch, let $B_g^c$ be the samples of group $g$ with label $c$, and let $\mathcal{Y}_{B_g}$ be the set of classes with at least two samples in $B_g$. For $c\in\mathcal{Y}_{B_g}$, we compute the sample mean and the unbiased coordinate-wise variances of the embeddings,
\begin{align*}
\hat m_g^c&=\frac{1}{|B_g^c|}\sum_{i\in B_g^c}\phi_g(x_i),\\
\hat v_{g,j}^c&=\frac{1}{|B_g^c|-1}\sum_{i\in B_g^c}\big(\phi_g(x_i)_j-\hat m_{g,j}^c\big)^2,\quad j=1,\dots,k.
\end{align*}
The minibatch alignment loss is the average over the eligible classes,
\[
\hat\Loss^{\mathrm{align}}_g=\frac{1}{|\mathcal{Y}_{B_g}|}\sum_{c\in\mathcal{Y}_{B_g}}\Wtwo^2\Big(\mathcal{N}\big(\hat m_g^c,\operatorname{diag}(\hat v_g^c)\big),\mu_c\Big),
\]
which is evaluated with \eqref{eq:w2diag}, and $\hat\Loss^{\mathrm{align}}_g=0$ if $\mathcal{Y}_{B_g}$ is empty. Algorithm~\ref{alg:train} uses this same eligible-class average.
\newline
The loss is differentiated with respect to the parameters that it depends on: the encoder $\phi_g$, through the embeddings $\phi_g(x_i)$ in the sample moments, and the anchors, through $m_c$ and $v_c$ in $\mu_c$. A gradient step on $\hat\Loss^{\mathrm{align}}_g$ therefore moves the embeddings of class $c$ in group $g$ toward the anchor of that class, and moves the anchor toward the embeddings of the groups. The head $\psi$ does not appear in this loss, so it is not updated by it.
\subsection{Minibatch Estimator of the Separation Loss with a Frozen Head}
We estimate the separation loss \eqref{eq:sep} with $M$ samples from each anchor, drawn with the reparameterization trick,
\[
\hat\Loss^{\mathrm{sep}}=\frac{1}{CM}\sum_{c=1}^C\sum_{r=1}^M\Loss\big(\mathrm{sg}(\psi);z_{c,r},c\big),\qquad z_{c,r}\sim\mu_c,
\]
with $M=4$ on NHANES and Fed-Heart and $M=16$ on EMBED. Algorithm~\ref{alg:train} displays the same $M$-sample estimator.
\newline
The stop-gradient $\mathrm{sg}$ means that the head is frozen when we differentiate this loss: its parameters are treated as constants. It is implemented by evaluating the head with detached parameters. The loss is then differentiated with respect to the anchors only, through the samples $z_{c,r}$, which depend on $m_c$ and $v_c$.
\newline
Since the samples of the separation loss come from the anchors and not from patients, we freeze the head. If this loss also updated the head, it would act as an additional training signal for the head on synthetic samples, for instance by making the head more confident on them, without necessarily moving the anchors. With the head frozen, the anchors move to regions that the head, trained on the patients, classifies correctly.

\section{Estimation of the References}
\label{app:floors}\label{app:refvalues}
This appendix gives the details of Algorithm~\ref{alg:floors}: how we implement the estimation of the references, and the guarantees that the references satisfy.

\subsection{Candidate Architectures and Their Training}
For each group, the set $\mathcal{A}$ contains $14$ candidates that differ in their architecture, in the patients that they are trained on, and in the loss that they optimize. The candidates are the following.
\begin{itemize}
    \item \emph{The constant predictor.} It outputs the class frequencies of the training patients for every patient, and it does not use the features.
    \item \emph{The joint model.} It has the form of the model of Section~\ref{sec:model}: an encoder $\phi_g$ for each group, with two layers and layer normalization into a latent space of dimension $64$, and a shared head $\psi$. It is trained without the alignment and separation losses, with equal weights for the groups, and with the class weights defined below:
    \[
    \min_{\phi_1,\dots,\phi_G,\,\psi}\ \sum_{g=1}^{G}\hat\Loss_g(\psi\circ\phi_g),
    \]
    where $\hat\Loss_g$ is the empirical task loss of group $g$ on its training patients.
    \item \emph{Flat predictors.} We try three architectures that map the features of group $g$ directly to class scores, without an encoder and a shared head: a linear model, and two different neural networks with two hidden layers of $32$ or $64$ units, each followed by a nonlinearity, with layer normalization after the first. Each architecture is trained either on the patients of group $g$ only, or on the patients of all the groups, using only the features of group $g$, which are set to zero for the patients who lack them. Finally, each is trained with or without inverse-frequency class weights. With these weights, the loss is $\sum_{i}w_{y_i}\Loss(f;x_i,y_i)\big/\sum_{i}w_{y_i}$ with $w_c\propto1/n_c$, where $n_c$ is the number of training patients of class $c$, so that patients of a rare class count more. This gives $3\times2\times2=12$ candidates.
\end{itemize}
All the candidates are trained by full-batch Adam, i.e., every step uses all the training patients of the candidate, with learning rate $10^{-2}$, and weight decay $10^{-4}$. We stop when the training loss has not improved by more than $10^{-5}$ for $60$ steps, or after $1500$ steps. The held-out loss that selects and scores the candidates is always the unweighted cross-entropy, which is the loss in which the references are used in Algorithm~\ref{alg:train}.

\subsection{Cross-Validation, Bootstrap Margin and Data Splits}

The training patients of each group are split into $K=5$ folds, stratified by label, and the $k$-th folds of all the groups are held out together (line~\ref{alg:fsplit}). For each group $g$ and each fold $k$, the candidates are compared by a stratified cross-validation with $K_{\mathrm{in}}=3$ folds on $\mathcal{D}^{\mathrm{in}}_{g,k}$, as described by the function \textsc{SelectPredictor}. The selected candidate is trained on all of $\mathcal{D}^{\mathrm{in}}_{g,k}$ and scored once on the held-out fold $S_{g,k}$, which gives the losses $\ell_{g,k}$. A group with fewer than six patients in $\mathcal{D}^{\mathrm{in}}_{g,k}$ uses the constant predictor.
\newline
The margin $c_g$ is computed with $N_{\mathrm{boot}}=4000$ bootstrap resamples and an error level $\beta=0.05$.
\newline
Since the model class contains constant predictors, the implementation also truncates the raw estimate at the held-out loss of the constant predictor. Let $\Const{g}$ be its mean on the same outer folds. The reference used in training is
\[
\RefPt{g}=\max\big\{\min\{\Est{g},\Const{g}\}-c_g,\,0\big\},
\]
and Equation~\eqref{eq:refpt} and Algorithm~\ref{alg:floors} omit the minimum with $\Const{g}$ for brevity. The quantity $\Const{g}$ estimates the loss of a predictor of the class, but it need not be an upper bound on $\Ach{g}$ in every sample.
\newline
The references only use the training patients of each group, and never the test patients. For Fed-Heart, where the experiments use outer folds, the references of each outer fold use only the patients outside its test fold, with the same fold assignment as the training runs, and the imputation statistics are fitted on these patients. For NHANES, the estimator uses the same split of the participants into training and test as the training runs, with the same seed, so the test participants coincide; the references use the training participants, including those that the training runs later hold out as validation, and never a test participant.
\newline
On EMBED, the references are estimated differently: for each group, the $5$-fold out-of-fold cross-entropy on the cached features of a model trained on that group alone and of a model trained jointly on all the groups, and the reference is the smaller of the two, without margin.
\newline
This procedure selects the smaller of two estimates that are computed on the same folds, and it has no margin, so the guarantee of Proposition~\ref{prop:estimate} does not cover it, and its references may be optimistic, mostly for the smallest groups.

\subsection{Error of the Raw Estimate}
\label{app:ref-estimate}

The references are estimates, so they have errors. In this subsection we show what error the raw estimate of Algorithm~\ref{alg:floors} has, and in the next one we show how the errors of the references affect the solution.
\newline
Let $\mathcal{C}_g$ be the set of predictors on $\Xcal_g$ that the candidates in $\mathcal{A}$ can represent, and let
\[
\Loss_g^{\mathcal{A}}:=\inf_{f\in\mathcal{C}_g}\Loss_g(f)
\]
be the best loss among them. If one of the candidates has the architecture of the model, as the joint model does, then $\Fcal_g\subseteq\mathcal{C}_g$ and $\Loss_g^{\mathcal{A}}\le\Ach{g}$.
\newline
The following proposition says that the raw estimate does not underestimate $\Loss_g^{\mathcal{A}}$: on average, it is not below the best loss that the candidates can reach. The reason is that a predictor that is trained on some patients and evaluated on others cannot, on average, do better than the best predictor of its class. It applies to the procedure of Algorithm~\ref{alg:floors}, i.e., to NHANES and Fed-Heart, and not to the simplified procedure of EMBED.

\begin{proposition}\label{prop:estimate}
Let $\hat f_{g,k}$ be the predictor trained in line~\ref{alg:ffit} of Algorithm~\ref{alg:floors} for group $g$ and outer fold $k$. Assume that:
\begin{enumerate}
    \item[(A1)] the patients of group $g$ are i.i.d. samples from $\Prob^g_{X,Y}$;
    \item[(A2)] the $K$ outer folds $S_{g,1},\dots,S_{g,K}$ have equal size, and the assignment of the patients to the folds does not depend on their observations;
    \item[(A3)] $\hat f_{g,k}$ is a function of the patients in $\mathcal{D}^{\mathrm{in}}_{g,k}$, of auxiliary training data that are independent of the patients in $S_{g,k}$, and of internal randomness that is independent of them. In particular, the selection of the architecture, the preprocessing and the training do not use the patients in $S_{g,k}$;
    \item[(A4)] $\hat f_{g,k}\in\mathcal{C}_g$.
\end{enumerate}
Then the raw estimate $\Est{g}$, before the truncation and the margin, satisfies
\[
\E\big[\Est{g}\big]\ge\Loss_g^{\mathcal{A}}.
\]
\end{proposition}
\begin{proof}
Since the loss is nonnegative, all the expectations below are well defined. We want to show that the held-out loss of every fold is at least $\Loss_g^{\mathcal{A}}$ in expectation.
\newline
Let $\hat R_{g,k}$ be the average loss of $\hat f_{g,k}$ on its held-out fold,
\[
\hat R_{g,k}=\frac{1}{|S_{g,k}|}\sum_{(x,y)\in S_{g,k}}\Loss(\hat f_{g,k};x,y).
\]
Since the folds have equal size, the average of all the held-out losses is the average of the fold averages, $\Est{g}=\frac1K\sum_{k=1}^K\hat R_{g,k}$. By linearity of the expectation, it is enough to show $\E[\hat R_{g,k}]\ge\Loss_g^{\mathcal{A}}$ for every $k$.
\newline
Fix a fold $k$, and let $\mathcal{I}_k$ be all the information that is used to produce $\hat f_{g,k}$: the patients in $\mathcal{D}^{\mathrm{in}}_{g,k}$, the auxiliary data, the internal randomness, and the fold assignment. Then $\hat f_{g,k}$ is fixed once $\mathcal{I}_k$ is given. By (A1)--(A3), conditional on $\mathcal{I}_k$, the patients in $S_{g,k}$ are i.i.d. samples from $\Prob^g_{X,Y}$. For a fixed predictor $f$, the average loss on such a sample has expectation $\Loss_g(f)$. Applying this to $f=\hat f_{g,k}$,
\[
\E\big[\hat R_{g,k}\mid\mathcal{I}_k\big]=\Loss_g(\hat f_{g,k}).
\]
By (A4), $\hat f_{g,k}\in\mathcal{C}_g$, so
\[
\Loss_g(\hat f_{g,k})\ge\inf_{f\in\mathcal{C}_g}\Loss_g(f)=\Loss_g^{\mathcal{A}}.
\]
Taking the expectation over $\mathcal{I}_k$ and then averaging over the folds,
\begin{align*}
\E\big[\Est{g}\big]
&=\frac1K\sum_{k=1}^K\E\Big[\E\big[\hat R_{g,k}\mid\mathcal{I}_k\big]\Big]\\
&=\frac1K\sum_{k=1}^K\E\big[\Loss_g(\hat f_{g,k})\big]\\
&\ge\Loss_g^{\mathcal{A}}.
\end{align*}
\end{proof}

The inner cross-validation is what keeps the selection of the architecture out of the scored fold, which assumption (A3) requires. If we scored the candidates on the same patients that select them, we would obtain $\E[\min_a\hat R_a]\le\min_a\E[\hat R_a]$ for the estimated risks $\hat R_a$ of the candidates, which is the optimistic selection effect. The outer folds of the implementation are stratified by label, so (A2) holds only approximately, and the proposition does not cover that procedure exactly. The proposition is also not a confidence bound: it gives the sign of the error of the raw estimate on average, and not its size.

\subsection{Effect of the Errors of the References}
\label{app:reference-stability}

The algorithm minimizes the worst-group excess loss computed with the references, $\max_g\big(\Loss_g(f_g)-\RefPt{g}\big)$, while what we want is the worst-group excess loss computed with the optimal losses, $\max_g\big(\Loss_g(f_g)-\Ach{g}\big)$. The two differ because the reference of group $g$ has an error $\delta_g=\RefPt{g}-\Ach{g}$. We show that the solution is affected by these errors only through their spread $s=\max_g\delta_g-\min_g\delta_g$. Intuitively, if all the references are shifted by the same constant, every excess loss is shifted by this constant, and no minimizer changes.
\newline
Let $\Fcal_{\mathrm{joint}}=\{(\psi\circ\phi_g)_{g=1}^{G}:\psi\in\Psi,\ \phi_g\in\Phi_g\}$ be the set of models in which all the groups share the head. The maximum over $\lambda\in\Simplex$ in \eqref{eq:excessdro} is attained at a vertex, so the problem that we solve is the minimization of $\max_g\big(\Loss_g(f_g)-\RefPt{g}\big)$ over $\Fcal_{\mathrm{joint}}$.

\begin{proposition}[Only the spread of the errors matters]\label{prop:spread}
Let $\hat f\in\Fcal_{\mathrm{joint}}$ be an approximate minimizer of the problem with the references,
\[
\max_g\big(\Loss_g(\hat f_g)-\RefPt{g}\big)\le\inf_{f\in\Fcal_{\mathrm{joint}}}\max_g\big(\Loss_g(f_g)-\RefPt{g}\big)+\varepsilon_{\mathrm{opt}}.
\]
Then
\[
\max_g\big(\Loss_g(\hat f_g)-\Ach{g}\big)\le\inf_{f\in\Fcal_{\mathrm{joint}}}\max_g\big(\Loss_g(f_g)-\Ach{g}\big)+s+\varepsilon_{\mathrm{opt}}.
\]
\end{proposition}
\begin{proof}
To shorten the notation, let $E(f)=\max_g\big(\Loss_g(f_g)-\Ach{g}\big)$ be the excess loss with the optimal losses and $\widehat E(f)=\max_g\big(\Loss_g(f_g)-\RefPt{g}\big)$ the one with the references, and let $\delta_{\max}$ and $\delta_{\min}$ be the largest and the smallest of the errors $\delta_g$, so that $s=\delta_{\max}-\delta_{\min}$.
\newline
We first compare $\widehat E$ with $E$. For every model $f$ and every group $g$,
\[
\Loss_g(f_g)-\RefPt{g}=\big(\Loss_g(f_g)-\Ach{g}\big)-\delta_g,
\]
and $\delta_{\min}\le\delta_g\le\delta_{\max}$, so
\[
\big(\Loss_g(f_g)-\Ach{g}\big)-\delta_{\max}\ \le\ \Loss_g(f_g)-\RefPt{g}\ \le\ \big(\Loss_g(f_g)-\Ach{g}\big)-\delta_{\min}.
\]
Taking the maximum over $g$ in the three terms,
\begin{equation}
E(f)-\delta_{\max}\ \le\ \widehat E(f)\ \le\ E(f)-\delta_{\min}\qquad\text{for every }f.
\label{eq:shift-bound}
\end{equation}
In this notation, the assumption of the proposition is $\widehat E(\hat f)\le\inf_{f'}\widehat E(f')+\varepsilon_{\mathrm{opt}}$, and we want to show that $E(\hat f)\le\inf_fE(f)+s+\varepsilon_{\mathrm{opt}}$. The idea is that $E$ and $\widehat E$ differ by an amount between $\delta_{\min}$ and $\delta_{\max}$, so a model that is nearly optimal for $\widehat E$ is nearly optimal for $E$, up to $\delta_{\max}-\delta_{\min}=s$.
\newline
Let $f\in\Fcal_{\mathrm{joint}}$ be any model. By the assumption on $\hat f$,
\[
\widehat E(\hat f)\le\inf_{f'}\widehat E(f')+\varepsilon_{\mathrm{opt}}\le\widehat E(f)+\varepsilon_{\mathrm{opt}}.
\]
Then
\begin{align*}
E(\hat f)&\le\widehat E(\hat f)+\delta_{\max} && \text{(left side of \eqref{eq:shift-bound}, for }\hat f)\\
&\le\widehat E(f)+\varepsilon_{\mathrm{opt}}+\delta_{\max} && \text{(optimality of }\hat f\text{ for }\widehat E)\\
&\le E(f)-\delta_{\min}+\varepsilon_{\mathrm{opt}}+\delta_{\max} && \text{(right side of \eqref{eq:shift-bound}, for }f)\\
&=E(f)+s+\varepsilon_{\mathrm{opt}}.
\end{align*}
This holds for every $f\in\Fcal_{\mathrm{joint}}$, so taking the infimum over $f$ on the right-hand side gives the result. The infimum does not need to be attained.
\end{proof}

The bound is an upper bound on the possible suboptimality, and it does not say that every nonzero spread causes a loss in performance. It also shows what Proposition~\ref{prop:estimate} does not need to give: the references do not have to be accurate, only equally inaccurate across groups.
\newline
We now extend the result to the full objective \eqref{eq:objective}, where the alignment loss is inside the maximum and the separation loss is added. Let $\theta=(\phi_1,\dots,\phi_G,\psi,\mu_1,\dots,\mu_C)\in\Theta$ be all the parameters, and write $\Loss_g(\theta)$ for $\Loss_g(\psi\circ\phi_g)$. With the maximum over $\lambda$ replaced by the maximum over the groups, the objective is
\[
J(\theta)=\max_g\Big\{\Loss_g(\theta)+\alpha_{\mathrm{align}}\Loss^{\mathrm{align}}_g(\theta)-\Ach{g}\Big\}+\alpha_{\mathrm{sep}}\Loss^{\mathrm{sep}}(\theta),
\]
and the problem that we solve replaces $\Ach{g}$ by $\RefPt{g}$ and the losses by their estimates,
\[
\widehat J(\theta)=\max_g\Big\{\widehat\Loss_g(\theta)+\alpha_{\mathrm{align}}\widehat\Loss^{\mathrm{align}}_g(\theta)-\RefPt{g}\Big\}+\alpha_{\mathrm{sep}}\widehat\Loss^{\mathrm{sep}}(\theta).
\]
We write $H_g(\theta)=\Loss_g(\theta)+\alpha_{\mathrm{align}}\Loss^{\mathrm{align}}_g(\theta)$ and $\widehat H_g(\theta)$ for its estimate, i.e., the terms inside the braces without the optimal loss or the reference.

\begin{proposition}[Stability of the full objective]\label{prop:regularized-spread}
Suppose that, on an event of interest, for every $\theta\in\Theta$ and every group $g$,
\[
\big|\widehat H_g(\theta)-H_g(\theta)\big|\le\eta_H,\qquad\big|\widehat\Loss^{\mathrm{sep}}(\theta)-\Loss^{\mathrm{sep}}(\theta)\big|\le\eta_S.
\]
If $\widehat J(\hat\theta)\le\inf_{\theta\in\Theta}\widehat J(\theta)+\varepsilon_{\mathrm{opt}}$, and the infima are finite, then on this event
\begin{equation}
J(\hat\theta)\le\inf_{\theta\in\Theta}J(\theta)+s+2\eta_H+2\alpha_{\mathrm{sep}}\eta_S+\varepsilon_{\mathrm{opt}}.
\label{eq:regularized-spread}
\end{equation}
\end{proposition}
\begin{proof}
The argument is that of Proposition~\ref{prop:spread}, with two additional differences between $\widehat J$ and $J$. For every group and every $\theta$,
\[
\widehat H_g(\theta)-\RefPt{g}=\big(H_g(\theta)-\Ach{g}\big)-\delta_g+\big(\widehat H_g(\theta)-H_g(\theta)\big),
\]
where $\delta_{\min}\le\delta_g\le\delta_{\max}$ and $|\widehat H_g(\theta)-H_g(\theta)|\le\eta_H$. Taking the maximum over $g$ and adding the separation terms, which differ by at most $\alpha_{\mathrm{sep}}\eta_S$, we obtain, for every $\theta$,
\[
J(\theta)-\delta_{\max}-d\ \le\ \widehat J(\theta)\ \le\ J(\theta)-\delta_{\min}+d,\qquad d:=\eta_H+\alpha_{\mathrm{sep}}\eta_S.
\]
Let $\theta\in\Theta$ be any comparison parameter. By the assumption on $\hat\theta$, $\widehat J(\hat\theta)\le\inf_{\theta'}\widehat J(\theta')+\varepsilon_{\mathrm{opt}}\le\widehat J(\theta)+\varepsilon_{\mathrm{opt}}$. Then
\begin{align*}
J(\hat\theta)&\le\widehat J(\hat\theta)+\delta_{\max}+d && \text{(left side of the bound, for }\hat\theta)\\
&\le\widehat J(\theta)+\varepsilon_{\mathrm{opt}}+\delta_{\max}+d && \text{(optimality of }\hat\theta\text{ for }\widehat J)\\
&\le J(\theta)-\delta_{\min}+d+\varepsilon_{\mathrm{opt}}+\delta_{\max}+d && \text{(right side of the bound, for }\theta)\\
&=J(\theta)+s+2d+\varepsilon_{\mathrm{opt}}.
\end{align*}
Taking the infimum over $\theta$, and using $2d=2\eta_H+2\alpha_{\mathrm{sep}}\eta_S$, gives \eqref{eq:regularized-spread}.
\end{proof}

Proposition~\ref{prop:spread} is the case of Proposition~\ref{prop:regularized-spread} with $\alpha_{\mathrm{align}}=\alpha_{\mathrm{sep}}=0$ and $\eta_H=\eta_S=0$. The proposition is a deterministic statement: it holds on any event on which the estimates are within $\eta_H$ and $\eta_S$ of the true losses, so the references may be computed from the same data as the training. It does not bound $s$, nor the probability of the event.
\newline
What we care about is the excess loss of the task, without the alignment and separation terms. Since these terms are nonnegative, it is at most $J(\hat\theta)$, and \eqref{eq:regularized-spread} gives, for every comparison parameter $\theta_0$,
\begin{align*}
\max_g\{\Loss_g(\hat\theta)-\Ach{g}\}&\le J(\theta_0)+s+2\eta_H\\
&\quad+2\alpha_{\mathrm{sep}}\eta_S+\varepsilon_{\mathrm{opt}}.
\end{align*}
The alignment and separation costs of $\theta_0$ are part of $J(\theta_0)$, so a comparison with a model that is optimal for the task alone includes the cost of aligning that model. The propositions assume that $\hat\theta$ is an approximate minimizer of $\widehat J$. Algorithm~\ref{alg:train} does not minimize $\widehat J$ (its head is not updated by the separation loss), and we do not prove that it reaches an approximate minimizer; the propositions describe how the errors of the references affect any such minimizer.

\section{Related Work}
\label{app:related}

\textbf{Heterogeneous feature spaces.} The simplest way to handle patients with different features is to build a common feature space. The first option is to keep only the features that all patients share, which discards the information that is available only in some groups \citep{rubin1976inference,schafer2002missing,little2019statistical}. The second option is to impute the missing features. MICE imputes each feature from the others with chained regressions \citep{vanbuuren2011mice}, GAIN trains a generative adversarial network to fill in the missing values \citep{yoon2018gain}, and ReMiND recovers missing neuroimaging with a diffusion model \citep{yuan2024remind}. Imputation can introduce error when the missing features are difficult to predict from the observed ones. Lipton et al. and Che et al. instead feed the missingness pattern of clinical time series to a recurrent network, since a test that was not ordered can be informative \citep{lipton2016directly,che2018recurrent}. Other works keep all the available information and train a model that tolerates missing modalities. SMIL addresses severely missing modalities in the training data \citep{ma2021smil}, and Ma et al. study how multimodal transformers degrade when modalities are missing \citep{ma2022robust}. ShaSpec models shared and modality-specific features \citep{wang2023shaspec}. FuseMoE and Flex-MoE use a mixture of experts to fuse an arbitrary combination of modalities \citep{han2024fusemoe,yun2024flexmoe}, and REMIND does so for medical data with missingness, with a focus on the combinations that are rare \citep{wu2026remind}. PRA-PoE uses a product of experts for the diagnosis of Alzheimer's disease with arbitrary missing modalities \citep{yang2026poe}. These methods can share information through common modalities and shared model components. Our framework explicitly aligns the representations of each class across groups through common anchors, including when the groups share no observed features.

\textbf{Shared representations.} Using a shared latent space to transfer knowledge is common in multitask and domain adaptation. Learning a representation shared by several tasks can reduce the number of samples that each task needs \citep{maurer2016benefit,du2021fewshot}, and matching the distributions of the representations of two domains can reduce the gap between them \citep{redko2017theoretical,shen2018wasserstein,madras2018learning}. Much of this work considers a common input space or a shared encoder. Class prototypes are also used in few-shot learning \citep{snell2017prototypical}, and in federated learning they have been used to align clients that have different models \citep{tan2022fedproto}. The closest work to our representation-learning component is FLIC, which learns personalised models for clients with different feature spaces by aligning their local embeddings with common Gaussian anchors \citep{rakotomamonjy2023flic}. We also organize the latent space around Gaussian class anchors, and we combine this alignment with a worst-group excess-loss objective and group-specific reference estimation.

\textbf{Group robustness.} Models trained to minimize the average loss give more weight to the larger groups. Standard Group DRO instead minimizes the loss of the worst-performing group \citep{sagawa2020distributionally,duchi2021learning}, and related work studies robustness for groups that are underrepresented in the data \citep{hashimoto2018fairness}. Generalization remains a challenge when the model is flexible enough to fit the training data \citep{sagawa2020distributionally}. Work on dataset shift and real-world benchmarks also motivates robustness to changes in the data distribution \citep{quinonero2009dataset,koh2021wilds}. While standard worst-group risk minimization compares groups by their raw loss, Oren et al. introduce a baselined loss that accounts for differences in the difficulty of language-modeling topics \citep{oren2019distributionally}. We adopt the minimax regret perspective \citep{savage1951theory,agarwal2022minimax} for heterogeneous feature spaces, because groups with different features can have different optimal losses (Proposition~\ref{prop:order}).

\textbf{Optimal losses and their estimation.} Several works measure the loss of each group against a reference. DoReMi optimizes the weights of the data domains of a language model with Group DRO, using the excess loss with respect to a reference model \citep{xie2023doremi}. Sheared LLaMA uses the same idea to decide how much data to load from each domain, with reference losses obtained from scaling laws or the evaluation losses of the source model \citep{xia2024sheared}. In our problem, each group has its own feature space, and we estimate its reference by evaluating candidate predictors on held-out patients from that group. The candidates include predictors trained on the group alone and predictors trained jointly across groups. For this we use cross-validation, which is the standard way to estimate the loss of a model on new patients. Choosing the model and evaluating it on the same patients makes the estimate optimistic, which nested cross-validation addresses by separating model selection from evaluation \citep{varma2006bias,tibshirani2009bias}. Note that cross-validation estimates the loss of a learning procedure, and not the infimum of the loss over the model class, which is the quantity that defines the excess loss. We therefore treat the estimates as references and not as exact values, and we study how the differences between their errors affect the solution (Appendix~\ref{app:floors}).

\end{document}